\documentclass[11pt,a4paper]{article}
\usepackage[hyperref]{emnlp-ijcnlp-2019}
\usepackage{times}
\usepackage{latexsym}

\usepackage{url}
\usepackage{graphicx}
\usepackage{hhline}
\usepackage{tcolorbox}
\usepackage{color}
\usepackage{amsmath}
\usepackage{amsfonts}
\usepackage{booktabs}
\usepackage{multirow}
\usepackage{makecell}
\usepackage{enumitem}
\usepackage{subfigure}
\usepackage{array}
\newcolumntype{L}[1]{>{\raggedright\let\newline\\\arraybackslash\hspace{0pt}}m{#1}}
\newcolumntype{C}[1]{>{\centering\let\newline\\\arraybackslash\hspace{0pt}}m{#1}}
\newcolumntype{R}[1]{>{\raggedleft\let\newline\\\arraybackslash\hspace{0pt}}m{#1}}
\newcolumntype{B}[1]{>{\raggedright\let\newline\\\arraybackslash\hspace{0pt}}p{#1}}
\newcolumntype{N}[1]{>{\centering\let\newline\\\arraybackslash\hspace{0pt}}p{#1}}
\newcolumntype{M}[1]{>{\raggedleft\let\newline\\\arraybackslash\hspace{0pt}}p{#1}}
\usepackage{pifont}
\newcommand{\cmark}{\ding{52}}%
\newcommand{\xmark}{\ding{56}}%

\DeclareMathOperator*{\argmax}{argmax\,}

\aclfinalcopy 

\title{Human-grounded Evaluations of Explanation Methods \\for Text Classification}

\author{Piyawat Lertvittayakumjorn {\normalfont and} Francesca Toni\\
  Department of Computing \\
  Imperial College London, UK\\
  \texttt{\{pl1515, ft\}@imperial.ac.uk}}

\date{}

\begin{document}
\maketitle
\begin{abstract}
Due to the black-box nature of deep learning models, methods for explaining the models' results are crucial to gain trust from humans and support collaboration between AIs and humans.  
In this paper, we consider several model-agnostic and model-specific explanation methods for CNNs for text classification 
and conduct three human-grounded evaluations, focusing  on different purposes of explanations:  \textit{(1)} revealing model behavior, \textit{(2)} justifying model predictions, and \textit{(3)} helping humans investigate uncertain predictions. 
The results highlight dissimilar qualities of the various explanation methods we consider and show the degree to which these methods could serve for each purpose.
\end{abstract}


\section{Introduction} \label{sec:int}
Explainable Artificial Intelligence (XAI) is aimed at providing explanations for decisions made by AI systems. 
The explanations are useful for supporting collaboration between AIs and humans in many cases \cite{Samek2018Overview}.
Firstly, if an AI outperforms humans in a certain task (e.g., AlphaGo \cite{AlphaGo}), humans can learn and distill knowledge from the given explanations. 
Secondly, if an AI's performance is close to human intelligence, the explanations can increase humans' confidence and trust in the AI \cite{symeonidis2009moviexplain}.
Lastly, if an AI is duller than humans, the explanations help humans verify the decisions made by the AI and also improve the AI \cite{biran2017human}.

One of the challenges in XAI is to explain prediction results given by deep learning models, which sacrifice transparency for high prediction performance. This type of explanations is called \textit{local explanations} as they explain individual predictions 
(in contrast to \textit{global explanations} which explain the trained model independently of any specific prediction). 
There have been several methods proposed to produce local explanations.
Some of them are model-agnostic, applicable to any machine learning model \cite{ribeiro2016lime,lundberg2017shap}. 
Others are applicable to a class of models such as neural networks \cite{lrpimage,sensitivityanalysis} or to a specific model such as Convolutional Neural Networks (CNNs) \cite{zhou2016cam}.

With so many explanation methods available, the next challenge is how to evaluate them so as to choose the right methods for different settings.
In this paper, we focus on human-grounded evaluations of local explanation methods for text classification. 
Particularly, we propose three evaluation tasks which target different purposes of explanations for text classification -- 
\textit{(1)} revealing model behavior to human users, 
\textit{(2)} justifying the predictions,
and
\textit{(3)} helping humans investigate uncertain predictions.
We then use the proposed tasks to evaluate nine explanation methods working on a standard CNN for text classification. These explanation methods are different in several aspects.
For example, regarding granularity, four explanation methods select words from the input text as explanations, whereas the other five select n-grams as explanations. 
In terms of generality, one of the explanation methods is model-agnostic, two are random baselines, another two (newly proposed in this paper) are specific to 1D CNNs for text classification, and the rest are applicable to neural networks in general.  
Overall, the contributions of our work can be summarized as follows.
\begin{itemize}
    \item We propose three human-grounded evaluation tasks to assess the quality of explanation methods with respect to different purposes of usage for text classification. (Section \ref{sec:hge})
    \item To increase diversity in the experiments, we develop two new explanation methods for CNNs for text classification. One is based on gradient-based analysis (\textbf{Grad-CAM-Text}). The other is based on model extraction using \textbf{decision trees}. (Section \ref{subsec:gradcam}-\ref{subsec:dt})
    \item We evaluate both new methods as well as random baselines and well-known existing methods using the three evaluation tasks proposed. The results highlight dissimilar qualities of the explanation methods and show the degree to which these methods could serve for each purpose. (Section \ref{sec:res})
\end{itemize}

\subsection{Terminology Used}
We use the following terms throughout the paper. 
\noindent (1) \textbf{Model}: a deep learning classifier we want to explain, e.g., a CNN.
\noindent (2) \textbf{Explanation}: an ordered list of text fragments (words or n-grams) in the input text which are most relevant to a prediction. Explanations for and against the predicted class are called \textbf{evidence} and \textbf{counter-evidence}, respectively. 
\noindent (3) \textbf{(Local) explanation method}: a method producing an explanation for a model and an input text.
\noindent (4) \textbf{Evaluation method}: a process to quantitatively assign to explanations scores which reflect the quality of the explanation method. 

\section{Background and Related Work} \label{sec:bgr}
This section discusses recent advances of explanation methods and evaluation for text classification as well as background knowledge about 1D CNNs -- the model used in the experiments. 

\subsection{Local Explanation Methods}
Generally, there are several ways to explain a result given by a deep learning model, such as explaining by examples \cite{Kim2014BCM} and generating textual explanations \cite{liu2018towards}. 
For text classification in particular, most of the existing explanation methods identify parts of the input text which contribute most towards the predicted class (so called \textit{attribution methods} or \textit{relevance methods}) by exploiting various techniques such as input perturbation \cite{Li2016Erasure}, gradient analysis \cite{sensitivityanalysis}, and relevance propagation \cite{ArrWASSA17}. 
Besides, there are other explanation methods designed for specific deep learning architectures such as attention mechanism \cite{attention-nli} and extractive rationale generation \cite{lei2016beer}.

We select some well-known explanation methods (which are applicable to CNNs for text classification) and evaluate them together with two new explanation methods proposed in this paper.

\subsection{Evaluation Methods}
Focusing on text classification, early works evaluated explanation methods by \textit{word deletion} -- gradually deleting words from the input text in the order of their relevance and checking how the prediction confidence drops \cite{arras-acl16, nguyen2018comparing}. 
\citet{arras2017relevant} and \citet{xiong2018looking} used relevance scores generated by explanation methods to construct document vectors by weighted-averaging word vectors and checked how well traditional machine learning techniques manage these document vectors.  
\citet{poerner2018evaluating} proposed two evaluation paradigms -- hybrid documents and morphosyntactic agreements. Both check whether an explanation method correctly points to the (known) root cause of the prediction.
Note that all of the aforementioned evaluation methods are conducted with no humans involved.

For human-grounded evaluation, \citet{mohseni2018human} proposed a benchmark which contains a list of relevant words for the actual class of each input text, identified by human experts. However, comparing human explanations with the explanations given by the tested method may be inappropriate since the mismatches could be due to not only the poor explanation method but also the inaccuracy of the model or the model reasoning differently from humans. 
\citet{nguyen2018comparing} asked humans to guess the output of a text classifier, given an input text with the highest relevant words highlighted by the tested explanation method. Informative (and discriminative) explanations will lead to humans' correct guesses. 
\citet{ribeiro2016lime} asked humans to choose a model which can generalize better by considering their local explanations. Also, they let humans remove irrelevant words, existing in the explanations, from the corpus to improve the prediction performance. 
Compared to previous work, our work is more comprehensive in terms of the various human-grounded evaluation tasks proposed and the number and dimensions of explanation methods being evaluated.   

\subsection{CNNs for Text Classification}
\begin{figure}[!t]
\centering
\includegraphics[width=0.48\textwidth]{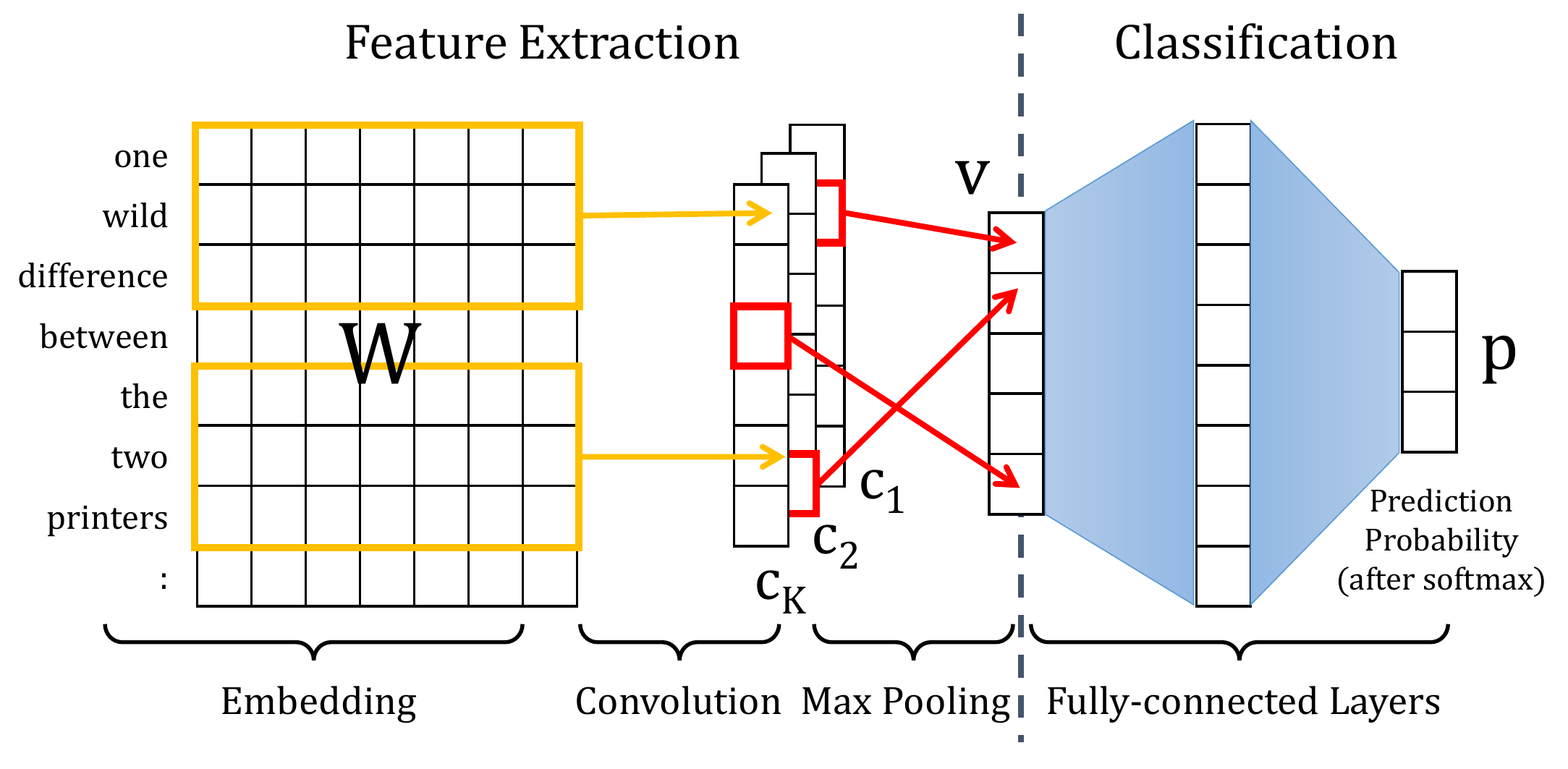}
\caption{CNN for text classification.}\label{fig:cnn}
\end{figure}

CNNs have been found to achieve promising results in many text classification tasks \cite{johnson2016cnnwordorder,gamback2017cnnhatespeech,zhangkumjornZeroShot}.
Figure \ref{fig:cnn} shows a standard 1D CNN for text classification which consists of four main steps: \textit{(i)} embedding an input text into an embedded matrix $\textbf{W}$; \textit{(ii)} applying $K$ fixed-size convolution filters to $\textbf{W}$ to find n-grams that possibly discriminate one class from the others; \textit{(iii)} pooling only the maximum value found by each filter, corresponding to the most relevant n-gram in text, to construct a filter-based feature vector, $\mathbf{v}$, of the input; and \textit{(iv)} using fully-connected layers ($FC$) to predict the results, and applying a softmax function to the outputs to obtain predicted probability of the classes ($\mathbf{p}$), i.e., $\mathbf{p} = softmax(FC(\mathbf{v}))$. 
While the original version of this model uses only one linear layer as $FC$ \cite{kim2014cnn}, more hidden layers can be added to increase the model capacity for prediction.  
Also, more than one filter size can be used to detect n-grams with short- and long-span relations \cite{conneau2017deepcnn}.   

\section{Human-grounded Evaluation Methods} \label{sec:hge}
We propose three human tasks to evaluate explanation methods for text classification as summarized in Table \ref{tab:evam}. Figure \ref{fig:aui} gives an example question for each task, discussed next.

\begin{table*}[ht!]
\centering
\small
\begin{tabular}{| C{0.11\linewidth} | L{0.26\linewidth} | L{0.26\linewidth} | L{0.26\linewidth} |} 
  \hline
  & \textbf{Task 1} (Section \ref{subsec:taska})& \textbf{Task 2} (Section \ref{subsec:taskb}) & \textbf{Task 3} (Section \ref{subsec:taskc})\\
  \hline
  Assumption & Good explanations can reveal model behavior
             & Good explanations justify the predictions
             & Good explanations help humans investigate uncertain predictions \\ \hline
  Model(s)  & Two classifiers with different performance on a test dataset 
            & One well-trained classifier
            & One well-trained classifier \\ \hline
  Input text    & A test example for which both classifiers predict the same class
                & A test example which the classifier predicts with high confidence ($\max_c\mathbf{p}_c > \tau_h$)
                & A test example which the classifier predicts with low confidence ($\max_c\mathbf{p}_c < \tau_l$)\\ \hline
  Information displayed & \makecell[l]{1. The input text \\ 
                                        2. The predicted class \\ 
                                        3. (Highlighted) top-$m$ evidence \\
                                        texts of each model
                                        } 
                & \makecell[l]{1. Top-$m$ evidence texts} 
                & \makecell[l]{1. The predicted class \\ 
                               2. The predicted probability $\mathbf{p}$\\
                               3. Top-$m$ evidence and \\
                               top-$m$ counter-evidence texts}  \\ \hline
  Human task & Select the more reasonable model and state if they are confident or not
            & Select the most likely class of the document which contains the evidence texts and state if they are confident or not
            & Select the most likely class of the input text and state if they are confident or not\\ \hline
  Scores to the explanation method & \makecell[l]{(-)1.0: (In)correct, confident \\ 
                        (-)0.5: (In)correct, unconfident \\ 
                        \quad 0.0: No preference}  
         & \makecell[l]{(-)1.0: (In)correct, confident \\ 
                        (-)0.5: (In)correct, unconfident \\ 
                        \quad 0.0: No preference}
         & \makecell[l]{(-)1.0: (In)correct, confident \\ 
                        (-)0.5: (In)correct, unconfident} \\ \hline
\end{tabular}
 \caption{A summary of the proposed human-grounded evaluation tasks.} \label{tab:evam}
\end{table*} 

\subsection{Revealing the Model Behavior} \label{subsec:taska}
Task 1 evaluates whether explanations can expose irrational behavior of a poor model. This property of explanation methods is very useful when we do not have a labelled dataset to evaluate the model quantitatively.   
To set up the task, firstly, we train two models to make them have different performance on classifying testing examples (i.e., different capability to generalize to unseen data). 
Then we use these models to classify an input text and apply the explanation method of interest to explain the predictions -- highlighting top-$m$ evidence text fragments on the text for each model. 
Next, we ask humans, based on the highlighted texts from the two models, which model is more reasonable? 
If the performance of the two models is clearly different, good explanation methods should enable humans to notice the poor model, which is more likely to decide based on non-discriminative words, even though both models predict the same class for an input text.

Additionally, there are some important points to note for this task. 
First, the chosen input texts must be classified into the same class by both models so the humans make decisions based only on the different explanations. However, it is worth to consider both the cases where both models correctly classify and where they misclassify. 
Second, we provide the choices for the two models along with confidence levels for the humans to select. If they select the right model with high confidence, the explanation method will get a higher positive score. In contrast, a confident but incorrect answer results in a large negative score. Also, the humans have the option to state no preference, for which the explanation method will get a zero score (See the last row of Table \ref{tab:evam}). 

\subsection{Justifying the Predictions} \label{subsec:taskb}
Explanations are sometimes used by humans as the reasons for the predicted class. This task tests whether the evidence texts are truly related to the predicted class and can distinguish it from the other classes, so called \textit{class-discriminative} \cite{selvaraju2017gradcam}. 
To set up the task, we use a well-trained model and select an input example classified by this model with high confidence ($\max_c\mathbf{p}_c > \tau_h$ where $\tau_h$ is a threshold parameter), so as to reduce the cases of unclear explanations due to low model accuracy or text ambiguity. (Note that we will look at low-confidence predictions later in Task 3.) Then we show only the top-$m$ evidence text fragments generated by the method of interest to humans and ask them 
to guess the class of the document containing the evidence.
The explanation method which makes the humans surely guess the class predicted by the model will get a high positive score. 
As in the previous task, this task considers both the correct and incorrect predictions with high confidence to see how well the explanations justify each of the cases.
For incorrect predictions, an explanation method gets a positive score when a human guesses the same incorrect class after seeing the explanation. In real applications, convincing explanations for incorrect classes can help humans understand the model's weakness and create additional fixing examples to retrain and improve the model.

\subsection{Investigating Uncertain Predictions} \label{subsec:taskc}
If an AI system makes a prediction with low confidence, it may need to raise the case with humans and let them decide, but with the analyzed results as additional information. This task aims to check if the explanations can help humans comprehend the situation and correctly classify the input text or not. 
To set up, we use a well-trained model and an input text classified by this model with low confidence ($\max_c\mathbf{p}_c < \tau_l$ where $\tau_l$ is a threshold parameter). 
Then we apply the explanation method of interest to find top-$m$ evidence and top-$m$ counter-evidence texts of the predicted class. 
We present both types of evidence to humans\footnote{We present counter-evidence as evidence for the other classes to simplify the task questions.} together with the predicted class and probability $\mathbf{p}$ and ask the humans to use all the information to guess the actual class of the input text, without seeing the input text itself. 
The scoring criteria of this task are similar to the previous tasks except that we do not provide the ``no preference'' option as the humans can still rely on the predicted scores when all the explanations are unhelpful.   

\section{Experimental Setup} \label{sec:exs}
\begin{figure}[!t]
\centering
\includegraphics[width=0.42\textwidth,height=17.7cm]{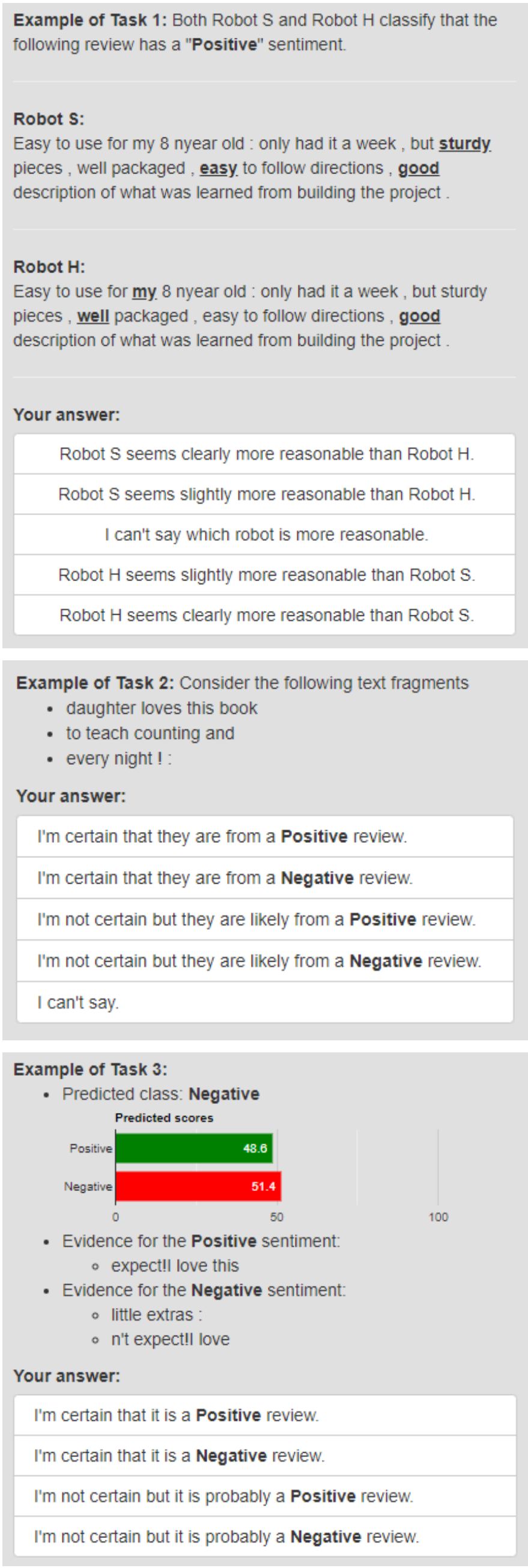}
\caption{Example questions and user interfaces. }\label{fig:aui}
\end{figure}

\subsection{Datasets}
We used two English textual datasets for the three tasks. 

\noindent (1) \textbf{Amazon} Review Polarity is a sentiment analysis dataset with positive and negative classes \cite{zhang2015dataset}. We randomly selected 100K, 50K, and 100K examples for training, validating, and testing the CNN models, respectively.

\noindent (2) \textbf{ArXiv} Abstract is a text classification dataset we created by collecting abstracts of scientific articles publicly available on ArXiv\footnote{https://arxiv.org}. 
Particularly, we collected abstracts from the ``Computer Science (CS)'', ``Mathematics (MA)'', and ``Physics (PH)'' categories, which are the three main categories on ArXiv. We then created a dataset with three disjoint classes removing the abstracts which belong to more than one of the three categories. 
In the experiments, we randomly selected 6K, 1.5K, and 10K examples for training, validating, and testing the CNN model, respectively. 

\subsection{Classification Models: 1D CNNs} \label{subsec:cnnsetup}
As for the classifiers, we used 1D CNNs with the same structures for all the tasks and datasets.
Specifically, we used 200-dim GloVe vectors
as non-trainable weights in the embedding layer \cite{pennington2014glove}. The convolution layer had three filter sizes [2, 3, 4] with 50 filters for each size, while the intermediate fully-connected layer had 150 units. The activation functions of the filters and the fully-connected layers are ReLU (except the softmax at the output layer). The models were implemented using Keras and trained with Adam optimizer. The macro-average F1 are 0.90 and 0.94 for the Amazon and the ArXiv datasets, respectively. 
Overall, the ArXiv appears to be an easier task as it is likely solvable by looking at individual keywords. 
In contrast, the Amazon sentiment analysis is not quite easy. Many reviews mention both pros and cons of the products, so a classifier needs to analyze several parts of the input to reach a conclusion. However, this is still manageable by the CNN architecture we used.

Also, in task 1, we need another model which performs worse than the well-trained model. 
In this experiment, we trained the second CNNs (i.e., the worse models) for the two datasets in different ways to examine the capability of explanation methods in two different scenarios. 
For the Amazon dataset, while the first (well-trained) CNN needed eight epochs until the validation loss converges, we trained the second CNN for only one epoch to make it underfitting. 
For the ArXiv dataset, we trained the second CNN using the same number of examples as the first model but with more specific topics. To explain, we randomly selected only examples from the subclass `Computation and Language', `Dynamical Systems', and `Quantum Physics' as training and validation examples for the class `Computer Science', `Mathematics', and `Physics', respectively. In other words, the training and testing data of the worse CNN came from different distributions. 
As a result, the macro-average F1 of the worse CNNs are 0.81 and 0.85 for the Amazon and the ArXiv datasets, respectively.

\subsection{Explanation Methods} \label{subsec:expm}
\begin{table}[t!]
\centering
\small
\begin{tabular}{|L{0.32\linewidth}|C{0.25\linewidth}|C{0.22\linewidth}|} 
 \hline
 \textbf{Method Name} & \textbf{Approach} & \textbf{Granularity} \\ \hline
  Random (W) & \multirow{2}{*}{\makecell[c]{Random\\Baselines}} & Words\\ \cline{1-1}\cline{3-3}
 Random (N) && N-grams \\ \hline
 LIME & Perturbation & Words \\ \hline
 LRP (W) & \multirow{4}{*}{\makecell[c]{Relevance\\Propagation}} & Words \\ \cline{1-1}\cline{3-3}
 LRP (N) && N-grams \\ \cline{1-1}\cline{3-3}
 DeepLIFT (W) && Words \\ \cline{1-1}\cline{3-3}
 DeepLIFT (N) && N-grams \\ \hline
 Grad-CAM-Text & Gradient & N-grams \\ \hline
 Decision Trees (DTs) & Model Extraction & N-grams \\
 \hline
\end{tabular}
\caption{Nine explanation methods evaluated.} \label{tab:expm}
\end{table} 

We evaluated nine explanation methods as summarized in Table \ref{tab:expm}. 
First, we used Random (W) and Random (N) as two baselines selecting words and non-overlapping n-grams randomly from the input text as evidence and counter-evidence. For the n-gram random baseline (and other n-gram based explanation methods in this paper), n is one of the CNN filter sizes [2, 3, 4].

Second, we selected LIME which is a well-known model-agnostic perturbation-based method \cite{ribeiro2016lime}. It trains a linear model using samples (5,000 samples in this paper) around the input text to explain the importance of each word towards the prediction. The importance scores can be either positive (for the predicted class) or negative (against the predicted class).

Third, we selected layer-wise relevance propagation (LRP), specifically $\epsilon$-LRP \cite{bach2015pixel}, and DeepLIFT \cite{shrikumar2017gradient} which are applicable to neural networks in general and performed very well in several evaluations using proxy tasks \cite{xiong2018looking, poerner2018evaluating}. LRP propagates the output of the target class (before softmax) back through layers to find attributing words, while DeepLIFT does the same but propagates the difference between the output and the predicted output of the reference input (i.e., all-zero embeddings in this paper). These two methods assign relevance scores to every word in the input text. Words with the highest and the lowest scores are selected as evidence for and counter-evidence against the predicted class, respectively.
Also, we extended LRP and DeepLIFT to generate explanations at an n-gram level. We considered all possible n-grams in the input text where n is one of the CNN filter sizes. Then the explanations are generated based on the relevance score of each n-gram, i.e., the sum of relevance scores of all words in the n-gram.

Next, we searched for model-specific explanation methods which target 1D CNNs for text classification. We found that \citet{jacovi2018understandcnn} proposed one: listing only n-grams corresponding to feature values in $\mathbf{v}$ (see Figure \ref{fig:cnn}) that pass thresholds for their filters. Each of the thresholds is set subject to sufficient purity of the classification results above it. However, their method is applicable to CNNs with only one linear layer as $FC$, while our CNNs have an additional hidden layer (with ReLU activation). So, we could not compare with their method in this work. To increase diversity in the experiments, we therefore propose two additional model-specific methods applicable to 1D CNNs with multiple layers in $FC$, presented next.

\begin{table*}[t!]
\setlength{\tabcolsep}{3pt}
\centering
\small
\begin{tabular}{|L{0.11\linewidth}|L{0.43\linewidth}|L{0.40\linewidth}|} 
 \multicolumn{3}{L{0.95\linewidth}}{\makecell[l]{
 \textbf{An example from the Amazon dataset, Actual: Pos, Predicted: Pos, (Predicted scores: Pos 0.514, Neg 0.486):} \\
 ``\textit{OK but not what I wanted: These would be ok but I didn't realize just how big they are. I wanted something I could } \\
 \textit{actually cook with. They are a full 12'' long. The handles didn't fit comfortably in
 my hand and the silicon tips are hard,} \\
 \textit{not rubbery texture like I'd imagined. The tips open to about 6'' between them.
 Hope this helps someone else know ...''}}
} \\
 \hline 
 \textbf{Method} & \textbf{Top-3 evidence texts} & \textbf{Top-3 counter-evidence texts} \\ \hline
 LIME (W) & comfortably / wanted / helps & not / else / someone\\ \hline
 LRP (W) & are / not / 6 & : / tips / open\\  \hline
 LRP (N) & \makecell[l]{are hard , not / about 6'' between / not what I wanted} & \makecell[l]{. The tips open / : These would / in my hand and}\\ \hline
 \makecell[l]{Grad-CAM-\\Text (N)} & \makecell[l]{comfortably in my hand / I wanted : These / \\. The tips open} & \makecell[l]{not what I wanted / not rubbery texture like / \\Hope this helps someone}\\ \hline
 DTs (N) & imagined . The tips & 'd imagined . / are . I wanted / would be ok\\
 \hline
\end{tabular}
 \caption{Examples of evidence and counter-evidence texts generated by some of the explanation methods.} \label{tab:example1}
\end{table*} 

\subsubsection{Grad-CAM-Text} \label{subsec:gradcam}
We adapt Grad-CAM \cite{selvaraju2017gradcam}, originally devised for explaining 2D CNNs, to find the most relevant n-grams for text classification. Since each value in the feature vector $\mathbf{v}$ corresponds to an n-gram selected by a filter, we use $E_{j,k}$ to show the effect of an n-gram selected by the $k$\textsuperscript{th} filter towards the prediction of class $j$:

\hspace{1.1cm} \( E_{j,k} = |\max(\frac{\partial FC(\mathbf{v})_j}{\partial \mathbf{v}_k}, 0)| \times \mathbf{v}_k.\) \quad (1)

\noindent The partial derivative term shows how much the prediction of class $j$ changes if the value from the $k$\textsuperscript{th} filter slightly changes. 
As we are finding the evidence for the target class $j$, we consider only the positive value of the derivative. 
Then $E_{j,k}$ combines this term with the strength of $\mathbf{v}_k$ to show the overall effect of the $k$\textsuperscript{th} filter for the input text. 
Next, we calculate the effect of each word $w_i$ in the input text by aggregating the effects of all the n-grams containing $w_i$. 

\hspace{1cm} \( E_{j,w_i} = \sum_{k}(E_{j,k}\times\mathbb{I}[w_i\in N_k])\)

\noindent where $N_k$ is an n-gram detected by the $k$\textsuperscript{th} filter.
Lastly, we select, as the evidence, non-overlapping n-grams which are detected by at least one of the filters and have the highest sums of the effects of all the words they contain. For example, to decide whether we will select the n-gram $N_k$ as an evidence text or not, we consider $\sum_{w_i\in N_k}E_{j,w_i}$.   
Note that we can find counter-evidence by changing, in equation (1), from $\max$ to $\min$.  

\subsubsection{Decision Trees} \label{subsec:dt}
This explanation method is based on model extraction \cite{bastani2017modelextraction}. We create a decision tree ($DT$) which mimics the behavior of the classification part (fully-connected layers) of the trained CNN. Given a filter-based feature vector $\mathbf{v}$, the DT needs to predict the same class as predicted by the CNN. Formally, we want 

\hspace{0.3cm}\(DT(\mathbf{v}) = \underset{j}\argmax \mathbf{p}_j = \underset{j}\argmax FC(\mathbf{v})_j.\)

\noindent For multi-class classification, we construct one DT for each class (one vs. rest classification). 
We employ CART with Gini index for learning DTs \cite{leo1984cart}. 
All the training examples are generated by the trained CNN using a training dataset, whereas a validation dataset is used to prune the DTs to prevent overfitting. 

Also, for each feature $v_j$ in $\mathbf{v}$, we calculate the Pearson's correlation between $v_j$ and the output of each class (before softmax) in $FC(\mathbf{v})$ using the training dataset, so we know which class is usually predicted given a high score of $v_j$ (i.e., correlated most to this feature). We use $c_j$ denoting the most correlated class of the feature $v_j$.

We can consider the DTs as a global explanation of the model as it explains the CNN in general.  
To create a local explanation, we use the DT of the predicted class to classify the input.  
At each decision node, we collect associated n-grams passing the nodes' thresholds to be evidence for (or counter-evidence against) the predicted class (depending on the most correlated class of each splitting feature). 
For example, an input text $X$ is classified to class $a$, so we use the DT of class $a$ to predict the input. If a decision node checks whether feature $v_j$ of this input is greater than 0.25 and assume it is true for this input, the n-gram corresponding to $v_j$ will be evidence if the most correlated class of $v_j$ is class $a$ (i.e., $c_j = a$). Otherwise, it will be counter-evidence if $c_j \neq a$.

\subsection{Implementations\footnote{The code and datasets of this paper are available at https://github.com/plkumjorn/CNNAnalysis}}
We used public libraries of LIME\footnote{https://github.com/marcotcr/lime}, LRP \cite{alber2018innvestigate}, and DeepLIFT\footnote{https://github.com/kundajelab/deeplift} in our experiments.
Besides, the code for computing Grad-CAM-Text was adapted from keras-vis\footnote{https://github.com/raghakot/keras-vis}, whereas we used scikit-learn \cite{pedregosa2011sklearn} for decision tree construction. All the DTs achieved over 80\% macro-F1 in mimicking the CNNs' predictions.

For the task parameters, we set $m$ = 3, $\tau_h=0.9$, and $\tau_l=0.7$. 
For each task and dataset, we used 100 input texts, half of which were classified correctly by the model(s) and the rest were misclassified. So, with nine explanation methods being evaluated, each task had 900 questions per dataset for human participants to answer. 
Examples of questions for each task are given in Figure \ref{fig:aui}.

For the Amazon dataset, we posted our tasks on Amazon Mechanical Turk (MTurk).
To ensure the quality of crowdsourcing, each question was answered by three workers and the scores were averaged.     
For the ArXiv dataset which requires background knowledge of the related subjects, we recruited graduates and post-graduate students in Computer Science, Mathematics, Physics, and Engineering to perform the tasks, and each question was answered by one participant. 
In total, we had 367 and 121 participants for the Amazon and the ArXiv datasets, respectively.

\section{Results and Discussion} \label{sec:res}
\begin{figure*}[!htb]
\centering
\includegraphics[width=\textwidth]{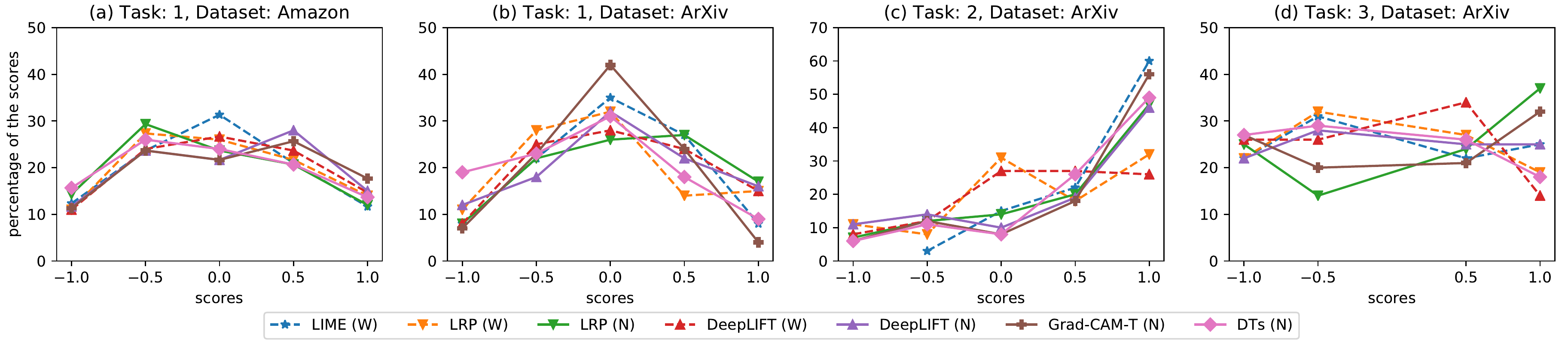}
\caption{Score distributions from task 1 of the Amazon dataset and from all the tasks of the ArXiv dataset.}\label{fig:scoreplots}
\end{figure*}

Examples of the generated explanations are shown in Table \ref{tab:example1} and a separate appendix.
Table \ref{tab:mainres} shows the average scores of each explanation method for each task and dataset, while Figure \ref{fig:scoreplots} displays the distributions of individual scores for all three tasks.
We do not show the distributions of tasks 2 and 3 of the Amazon dataset as they look similar to the associated ones of the ArXiv dataset.

\begin{table*}[t!]
\setlength{\tabcolsep}{2.4pt}
\centering
\small
\begin{tabular}{L{0.15\linewidth} |R{0.035\linewidth} R{0.035\linewidth} R{0.035\linewidth}|R{0.035\linewidth} R{0.035\linewidth} R{0.035\linewidth}||R{0.035\linewidth} R{0.035\linewidth} R{0.035\linewidth}|R{0.035\linewidth} R{0.035\linewidth} R{0.035\linewidth}||R{0.035\linewidth} R{0.035\linewidth} R{0.035\linewidth}|R{0.035\linewidth} R{0.035\linewidth} R{0.035\linewidth}} 
 \hline
 \multirow{3}{*}{\makecell[c]{Explanation \\ Method}} & \multicolumn{6}{c||}{\textbf{Task 1}}& \multicolumn{6}{c||}{\textbf{Task 2}}& \multicolumn{6}{c}{\textbf{Task 3}}\\ \cline{2-19}
  & \multicolumn{3}{c|}{Amazon}& \multicolumn{3}{c||}{ArXiv}& \multicolumn{3}{c|}{Amazon}& \multicolumn{3}{c||}{ArXiv}& \multicolumn{3}{c|}{Amazon}& \multicolumn{3}{c}{ArXiv}\\ 
  & $\mathcal{A}$ & \cmark & \xmark & $\mathcal{A}$ & \cmark & \xmark & $\mathcal{A}$ & \cmark & \xmark & $\mathcal{A}$ & \cmark & \xmark & $\mathcal{A}$ & \cmark & \xmark & $\mathcal{A}$ & \cmark & \xmark \\ \hline
  Random (W)& \underline{.02} & \underline{.00} & \underline{.04}
            &-.11 &-.05 &\underline{-.17} 
            & .06 & .10 & .02 
            &.07 &.09 &.04 
            & \underline{.05} & \underline{.53} & \underline{-.43}
            &\underline{.01} &.32 &\underline{-.30} \\
  Random (N)& \underline{.02} & \underline{.02} & \underline{.02}
            &-.12 &-.16 &\underline{-.07} 
            & .12 & .13 & .12
            &.29 &.32 &.25 
            & \underline{-.01} & \underline{.54} & \underline{-.55}
            &\underline{.02} &.29 &\underline{\textbf{-.25}} \\
  LIME (W)& \underline{-.02} & \underline{.02} & \underline{-.06}
            &\underline{.03} &\underline{.02} &\underline{\textbf{.03}} 
            & \underline{\textbf{.69}} & \underline{\textbf{.74}} & \underline{.64}
            &\underline{\textbf{.70}} &\underline{\textbf{.75}} &\underline{\textbf{.64}} 
            & \underline{.02} & \underline{.50} & \underline{-.45}
            &\underline{-.02} &.31 &\underline{-.34} \\
  LRP (W) & \underline{.00} & \underline{-.01} & \underline{.02}
            &\underline{-.03} &-.01 &\underline{-.05} 
            & .13 & .26 & -.01
            &.26 &.36 &.16 
            & \underline{-.02} & \underline{.50} & \underline{-.54} 
            &-.06 &.33 &\underline{-.44} \\
  LRP (N) & -.07 & \underline{-.04} & \underline{-.09} 
            &\underline{\textbf{.12}} &\underline{\textbf{.24}} &\underline{-.01} 
            & .26 & .45 & .08
            &.44 &.49 &.39 
            & \underline{.08} & \underline{.60} & \underline{-.43}
            &\underline{\textbf{.17}} &\underline{\textbf{.60}} &\underline{-.26} \\
  DeepLIFT (W)& \underline{.04} & \underline{.03} & \underline{.04}
            &\underline{.07} &\underline{.13} & \underline{.00} 
            & .21 & .37 & .04
            &.26 &.35 &.16 
            & \underline{-.03} & \underline{.47} & \underline{-.53}
            &-.08 &.28 &\underline{-.44} \\
  DeepLIFT (N)& \underline{.06} & \underline{.06} & \underline{\textbf{.05}}
            &\underline{.06} &\underline{.22} &\underline{-.10} 
            & .23 & .47 & -.01
            &.38 &.47 &.28 
            & \underline{.05} & \underline{.59} & \underline{-.49}
            &\underline{.02} &.33 &\underline{-.30} \\
  Grad-CAM-T (N)& \underline{\textbf{.07}} & \underline{\textbf{.11}} & \underline{.03}
            &\underline{-.03} &-.04 &\underline{-.01} 
            & \underline{.65} & \underline{.64} & \underline{\textbf{.66}}
            &.53 &\underline{.65} &\underline{.41} 
            & \underline{.05} & \underline{.51} & \underline{-.42}
            &\underline{.06} &\underline{.56} &\underline{-.45} \\
  DTs (N) & -.05 & \underline{-.02} & \underline{-.08}
            &-.13 &-.22 &\underline{-.03} 
            & \underline{.64} & \underline{.68} & \underline{.59}
            &.51 &\underline{.69} &.32 
            & \underline{\textbf{.10}} & \underline{\textbf{.60}} & \underline{\textbf{-.40}}
            &-.11 &.29 &-.50 \\ \hline
 Fleiss $\kappa$ (Amazon) & \multicolumn{3}{c|}{0.050 / 0.054} & \multicolumn{3}{c||}{N/A} & \multicolumn{3}{c|}{0.274 / 0.371} & \multicolumn{3}{c||}{N/A} & \multicolumn{3}{c|}{0.212 / 0.499} & \multicolumn{3}{c}{N/A} \\ \hline
\end{tabular}
\caption{The average scores of the three evaluation tasks. 
The score range is [-1,1] in which 1 is better. 
$\mathcal{A}$, \cmark, and \xmark  are for all, correctly classified, and misclassified input texts, respectively. 
Boldface numbers are the highest average scores in the columns. 
A number is underlined when there is no statistically significant difference between the scores of the corresponding method and the best method in the same column (at a significance level of 0.05).
The last row reports inter-rater agreement measures (Fleiss' kappa) in the format of $\alpha$ / $\beta$ where $\alpha$ considers answers with human confidence levels (5 categories for task 1-2 and 4 categories for task 3) and $\beta$ considers answers regardless of the human confidence levels (3 categories for task 1-2 and 2 categories for task 3).} \label{tab:mainres}
\end{table*} 

\subsection{Task 1}
For the Amazon dataset, though Grad-CAM-Text achieved the highest overall score, the performance was not significantly different from other methods including the random baselines. Also, the inter-rater agreement for this task was quite poor. It suggests that existing explanation methods cannot apparently reveal irrational behavior of the underfitting CNN to lay human users. So, the scores of most explanation methods distribute symmetrically around zero, as shown in Figure \ref{fig:scoreplots}(a). 

For the ArXiv dataset, LRP (N) and DeepLIFT (N) got the highest scores when both CNNs predicted correctly. Hence, they can help humans identify the poor model to some extent. However, there was no clear winner when both CNNs predicted wrongly. One plausible reason is that evidence for an incorrect prediction, even by a well-trained CNN, is usually not convincing unless we set a (high) lower bound of the confidence of the predictions (as we did in task 2).

Additionally, we found that psychological factors fairly affect this task. Based on the results, for two explanations with comparable semantic quality, humans prefer the explanation with more evidence texts and select it as more reasonable. This is consistent with the findings by \citet{zemla2017evaluating}. 
Conversely, the DTs method performed in the opposite way. The DT of the better model usually focuses on a few most relevant texts in the input and outputs fewer evidence texts. This possibly causes the low performance of the DTs method in this task. 
Also, we got feedback from the participants that they sometimes penalized an evidence text which is highlighted in a strange way, such as ``... greedy \underline{algorithm. In this} paper, we ...''. Hence, in real applications, syntax integrity should be taken into account to generate explanations. 

\subsection{Task 2}
LIME clearly achieved the best results in task 2 followed by Grad-CAM-Text and DTs. 
These methods are class discriminative, being able to find good evidence for the predicted class regardless of whether the prediction is correct. 

We believe that LIME performed well because it tests that the missing of evidence words from the input text greatly reduces the probability of the predicted class, so these words are semantically related to the predicted class (given that the model is accurate). 
Meanwhile, the DTs method selects evidence based on the most correlated class of the splitting features. So, the evidence n-grams are more likely related to the predicted class than the other classes. However, they may be less relevant than LIME's as the evidence is generated from a global explanation of the model (DTs).  
Besides, Grad-CAM-Text worked relatively well here probably because it preserves the class discriminative property of Grad-CAM \cite{selvaraju2017gradcam}.

By contrast, LRP and DeepLIFT generated acceptable evidence only for the correct predictions. 
Also, LRP (N) and DeepLIFT (N) performed better than LRP (W) and DeepLIFT (W) in both datasets. This might be because one evidence n-gram contains more information than one evidence word. Nevertheless, even the Random (N) method surpasses the LRP (W) and the DeepLIFT (W) for the ArXiv dataset. Thereby, whenever we use LRP and DeepLIFT, we should present to humans the most relevant words together with their contexts. 

\subsection{Task 3}
The negative scores under the \xmark columns of task 3 show that using explanations to rectify the predictions is not easy.
Hence, the overall average scores of many explanation methods stay close to zero.

DTs performed well only on the Amazon dataset. The average numbers of n-grams per explanation, generated by the DTs, are 2.00 and 1.77 for the Amazon and ArXiv datasets, respectively. Also, the reported n-grams could be repetitive and overlapping. This reduced the amount of useful information displayed, and it may be insufficient for humans to choose one of the CS, MA, and PH categories, which are more similar to one another than the positive and negative sentiments.

Meanwhile, LRP (N) performed consistently well on both datasets. This is reasonable considering our discussions in task 2. First, LRP (N) generates good evidence for correct predictions, so it can gain high scores in the \cmark columns. On the other hand, the evidence for incorrect predictions (\xmark) is usually not convincing, so the counter-evidence (which is likely to be the evidence of the correct class) can attract humans' attention. Furthermore, the fact that LRP is not class discriminative does not harm it in this task as humans can recognize an evidence text even if it is selected by the LRP (N) as counter-evidence (and vice versa).

For example, in the ArXiv dataset, we found a case in which the predicted class is PH (score = 0.48) but the correct class is CS (score = 0.07). LRP (N) selected `armed bandit settings with', `the Wasserstein distance', and `derive policy gradients' as evidence for the class PH. 
These n-grams, however, are not truly related to PH. Rather, they revealed the true class of this text and made a human choose the CS option with high confidence despite the low predicted score.

Regarding LIME, the situation is reversed as LIME can find both good evidence and counter-evidence. These make humans be indecisive and, possibly, select a wrong option as the explanation is presented at a word level (without any contexts).

\subsection{Model Complexity}
Apart from the results of the three tasks, it is worth to discuss the size of the DTs which mimic the four CNNs in our experiments. As shown in Table \ref{tab:dtsmt}, the size of the DTs can reflect the complexity of the CNNs. Although the well-trained CNN of the Amazon dataset got 0.9 F1 score, the DTs of this CNN needed more than 5,500 nodes to achieve 85\% fidelity (compared to only hundreds of nodes required for the ArXiv dataset). This illustrates the high complexity of the Amazon task compared to the ArXiv task even though both tasks were managed effectively by the same CNN architecture. 

For the ArXiv dataset, the DTs of the poor CNN are smaller than the ones of the well-trained CNN. This is likely because the poor CNN was trained on a specific dataset (i.e., selected subtopics of the main categories), so it had to deal with fewer discriminative patterns in texts compared to the first CNN trained using texts from all subtopics.


Further studies of this quality of the DTs would be useful for some applications, e.g., measuring model complexity \cite{2014-Bianchini-Complexity} and model compression \cite{2018-Cheng-Compression}.
    
\begin{table}[t!]
\setlength{\tabcolsep}{2.4pt}
\centering
\small
\begin{tabular}{L{0.37\linewidth} R{0.16\linewidth} R{0.16\linewidth} R{0.16\linewidth} }
\hline
    & \textbf{\#Nodes} &   \textbf{Depth} & \textbf{\#Leaves} \\
 \hline
    \multicolumn{4}{l}{\textbf{Amazon: 1\textsuperscript{st} CNN (well-trained)} -- \textit{F} = 0.85} \\
    Negative & 5535 & 38 & 2768 \\
    Positive & 5537 & 45 & 2769 \\

   \hline
   \multicolumn{4}{l}{\textbf{Amazon: 2\textsuperscript{nd} CNN (underfitting)} -- \textit{F} = 0.82} \\ 
   Negative & 6405 & 40 & 3203 \\
   Positive & 6369 & 40 & 3185 \\
   \hline
\multicolumn{4}{l}{\textbf{ArXiv: 1\textsuperscript{st} CNN (well-trained)} -- \textit{F} = 0.89} \\ 
    Computer Science& 363& 25& 182\\
Mathematics& 565& 24& 283\\
Physics& 325& 24& 163\\
   \hline
   \multicolumn{4}{l}{\textbf{ArXiv: 2\textsuperscript{nd} CNN (specific data)} -- \textit{F} = 0.84} \\
   Computer Science& 107& 17& 54\\
Mathematics& 263& 28& 132\\
Physics& 237& 29& 119\\

\hline
\end{tabular}
\caption{Metadata of the DTs in the experiments. \textit{F} refers to fidelity of the DTs (macro-average F1).} \label{tab:dtsmt}
\end{table}

\section{Conclusion} \label{sec:con}
We proposed three human tasks to evaluate local explanation methods for text classification. 
Using the tasks in this paper, we experimented on 1D CNNs and found that 
\textit{(i)} LIME is the most class discriminative method, justifying predictions with relevant evidence; 
\textit{(ii)} LRP (N) works fairly well in helping humans investigate uncertain predictions; 
\textit{(iii)} using explanations to reveal model behavior is challenging, and none of the methods achieved impressive results;
\textit{(iv)} whenever using LRP and DeepLIFT, we should present to humans the most relevant words together with their contexts
and \textit{(v)} the size of the DTs can also reflect the model complexity.
Lastly, we consider evaluating on other datasets and other advanced architectures beneficial future work as it may reveal further interesting qualities of the explanation methods. 

\section*{Acknowledgments}
We would like to thank all the participants in the experiments as well as Alon Jacovi and anonymous reviewers for helpful comments. 
Also, the first author would like to thank the support from Anandamahidol Foundation, Thailand. 

\bibliography{emnlp-ijcnlp-2019}
\bibliographystyle{acl_natbib}

\newpage
\appendix
\section{CNN models}
\label{sec:perfcnns}
This section reports the performance of the trained CNN models on a test set of each dataset.
\subsection{Amazon Dataset}
\begin{table}[h!]
\setlength{\tabcolsep}{2.4pt}
\centering
\small
\begin{tabular}{L{0.30\linewidth} R{0.15\linewidth} R{0.15\linewidth} R{0.13\linewidth} R{0.17\linewidth}} 
 \hline\hline
    \textbf{1\textsuperscript{st} CNN (better)} & Prec. &   Recall & F1 &  Support \\ \hline

    Negative  &     0.92&      0.89&      0.90&     50039 \\
    Positive  &     0.89&      0.92&      0.90&     49961 \\ \hline

   micro avg&       0.90&     0.90&      0.90&   100000 \\
   macro avg&       0.90&      0.90&      0.90&    100000 \\

 \hline\hline
 \textbf{2\textsuperscript{nd} CNN (worse)} & Prec. &   Recall & F1 &  Support \\ \hline

    Negative&       0.82 &     0.81 &     0.81  &   50039 \\
    Positive&       0.81 &     0.82 &     0.81  &   49961 \\ \hline

   micro avg&       0.81 &     0.81 &     0.81 &   100000 \\
   macro avg&       0.81 &     0.81 &     0.81 &   100000 \\
\hline\hline
\end{tabular}
\caption{Precision, Recall, and F1 scores of both CNNs for the Amazon dataset} \label{tab:cnnsamazon}
\end{table} 

\subsection{ArXiv Dataset}
\begin{table}[h!]
\setlength{\tabcolsep}{2.4pt}
\centering
\small
\begin{tabular}{L{0.30\linewidth} R{0.15\linewidth} R{0.15\linewidth} R{0.13\linewidth} R{0.17\linewidth}} 
 \hline\hline
    \textbf{1\textsuperscript{st} CNN (better)} & Prec. &   Recall & F1 &  Support \\ \hline

         Computer science&       0.94 &     0.93 &     0.93 &    10000 \\
        Mathematics&       0.92 &     0.93 &     0.92  &   10000 \\
        Physics&       0.96 &     0.94 &     0.95 &    10000 \\
\hline
   micro avg&       0.94 &     0.94 &     0.94 &    30000 \\
   macro avg&       0.94 &     0.94 &     0.94 &    30000 \\

 \hline\hline
 \textbf{2\textsuperscript{nd} CNN (worse)} & Prec. &   Recall & F1 &  Support \\ \hline

        Computer science  &     0.96 &     0.74  &    0.84 &    10000\\
        Mathematics&       0.75&      0.94&      0.83 &    10000\\
        Physics&       0.89 &     0.88&      0.89 &    10000\\
\hline
   micro avg&       0.85&      0.85&      0.85 &    30000\\
   macro avg&       0.87&      0.85&      0.85 &    30000\\

\hline\hline
\end{tabular}
\caption{Precision, Recall, and F1 scores of both CNNs for the ArXiv dataset} \label{tab:cnnsarxiv}
\end{table} 

\newpage
\section{Decision Trees}
This section reports the decision trees’ performance in mimicking the CNNs' predictions (i.e., fidelity) on the test sets. 
All the DTs achieved over 80\% macro-F1 in mimicking the CNNs’ predictions.
As the F1 scores say, it's easier for the decision trees to mimic the behavior of the well-trained CNNs than the poor CNNs.
\subsection{Amazon Dataset}
\begin{table}[h!]
\setlength{\tabcolsep}{2.4pt}
\centering
\small
\begin{tabular}{L{0.30\linewidth} R{0.15\linewidth} R{0.15\linewidth} R{0.13\linewidth} R{0.17\linewidth}} 
 \hline\hline
    \textbf{1\textsuperscript{st} CNN (better)} & Prec. &   Recall & F1 &  Support \\ \hline
    Negative&       0.84 &     0.84 &    0.84&     48333\\
    Positive&       0.85 &     0.85 &     0.85 &    51667\\
\hline
   micro avg  &     0.85 &     0.85 &     0.85&    100000\\
   macro avg  &     0.85 &     0.85 &     0.85&    100000\\
   \hline\hline
   \textbf{2\textsuperscript{nd} CNN (worse)} & Prec. &   Recall & F1 &  Support \\ \hline
   Negative &      0.81 &     0.82&      0.82 &    49482\\
   Positive &      0.82 &     0.82  &    0.82 &    50518\\
\hline
   micro avg &      0.82  &    0.82 &     0.82 &   100000\\
   macro avg &      0.82 &     0.82 &     0.82 &   100000\\
\hline\hline
\end{tabular}
\caption{Performance of the decision trees in mimicking the CNNs' predictions for the Amazon dataset} \label{tab:dtsamazon}
\end{table} 

\subsection{ArXiv Dataset}
\begin{table}[h!]
\setlength{\tabcolsep}{2.4pt}
\centering
\small
\begin{tabular}{L{0.30\linewidth} R{0.15\linewidth} R{0.15\linewidth} R{0.13\linewidth} R{0.17\linewidth}} 
 \hline\hline
    \textbf{1\textsuperscript{st} CNN (better)} & Prec. &   Recall & F1 &  Support \\ \hline
Computer science  &     0.89   &   0.91  &    0.90  &    9971\\
Mathematics    &   0.89   &   0.87   &   0.88  &   10203\\
Physics    &   0.90   &   0.91  &    0.90  &    9826\\
\hline
   micro avg   &    0.89  &    0.89   &   0.89  &   30000\\
   macro avg   &    0.89  &    0.89   &   0.89  &   30000\\

   \hline\hline
   \textbf{2\textsuperscript{nd} CNN (worse)} & Prec. &   Recall & F1 &  Support \\ \hline
Computer science   &    0.83  &    0.81   &   0.82   &   7653 \\
       Mathematics &      0.82 &     0.88  &    0.85  &   12506\\
       Physics     &  0.88   &   0.81  &    0.84  &    9841\\
\hline
   micro avg     &  0.84 &     0.84  &    0.84   &  30000\\
   macro avg     &  0.84 &     0.83   &   0.84   &  30000\\

\hline\hline
\end{tabular}
\caption{Performance of the decision trees in mimicking the CNNs' predictions for the ArXiv dataset} \label{tab:dtsarxiv}
\end{table} 

\newpage
\section{Examples of the Explanations}

\noindent \underline{\textbf{Example 1}}: Amazon Dataset, (Actual: Positive, Predicted: Negative)

\vspace{0.2cm}
\noindent\fbox{%
    \parbox{0.97\linewidth}{%
        ``Source hip hop hits Volume 3: THe songs listed aren't even on the CD! I bought it for Bling Bling and it wasn't on the CD. the other songs are good, but not what I was looking for. Amazon needs to get the info right on this listing.''
    }%
}

\vspace{0.2cm}
\noindent\textbf{Top-5 evidence texts}
\begin{itemize}
    \item Random (W): . / get / hip / was / I
    \item Random (N):	the CD ! I / the CD / needs to get the / info right on this / for .
    \item LIME:	not / bought / 3 / info / Bling
    \item LRP (W):	it / bought / . / listed / :
    \item LRP (N):	! I bought it / : THe songs listed / was looking for . / right on this listing / not what I
    \item DeepLIFT (W):	it / bought / . / listed / :
    \item DeepLIFT (N):	! I bought it / : THe songs listed / was looking for . / right on this listing / not what I
    \item Grad-CAM-Text:	n't even on the / not what I was / hits Volume 3 : / CD ! I / . Amazon needs to
    \item DTs:	n't even on the / CD ! I
\end{itemize}

\newpage
\noindent \underline{\textbf{Example 2}}: ArXiv Dataset, (Actual: Physics (PH), Predicted: Computer Science (CS))

\vspace{0.2cm}
\noindent\fbox{%
    \parbox{0.97\linewidth}{%
        ``Multiple-valued Logic (MVL) circuits are one of the most attractive applications of the Monostable-to-Multistable transition Logic (MML), and they are on the basis of advanced circuits for communications. The operation of such quantizer has two steps : sampling and holding. Once the quantizer samples the signal, it must maintain the sampled value even if the input changes. However, holding property is not inherent to MML circuit topologies. This paper analyses the case of an MML ternary inverter used as a quantizer, and determines the relations that circuit representative parameters must verify to avoid this malfunction.''
    }%
}

\vspace{0.2cm}
\noindent\textbf{Top-5 evidence texts}
\begin{itemize}
    \item Random (W):	not / This / one / basis / MML
    \item Random (N):	) , and / , holding property is / are one of / sampled value even / circuit topologies
    \item LIME:	paper / Logic / circuits / communications / applications
    \item LRP (W):	paper / - / communications / topologies / the
    \item LRP (N):	topologies . This paper / to - Multistable transition / valued Logic ( MVL / circuits for communications . / the quantizer samples the
    \item DeepLIFT (W):	paper / - / communications / Logic / the
    \item DeepLIFT (N):	topologies . This paper / valued Logic ( MVL / to - Multistable transition / circuits for communications . / the quantizer samples the
    \item Grad-CAM-Text:	circuits for communications . / ( MVL ) circuits / MML ternary inverter used / topologies . This paper / - valued Logic
    \item DTs:	MML ternary inverter / MVL ) circuits are / advanced circuits / circuits for communications / to avoid this malfunction
\end{itemize}

\newpage
\noindent \underline{\textbf{Example 3}}: Amazon Dataset, (Actual: Positive, Predicted: Positive),
Predicted scores: Positive (0.514), Negative (0.486)

\vspace{0.2cm}
\noindent\fbox{%
    \parbox{0.97\linewidth}{%
``OK but not what I wanted: These would be ok but I didn't realize just how big they are. I wanted something I could actually cook with. They are a full 12'' long. The handles didn't fit comfortably in my hand and the silicon tips are hard, not rubbery texture like I'd imagined. The tips open to about 6'' between them.Hope this helps someone else know better if it's what they want.''
    }%
}

\vspace{0.2cm}
\noindent\textbf{Top-5 evidence texts}
\begin{itemize}
        \item Random (W):	not / wanted / 'd / with / The
    \item Random (N):	did n't / be ok / could actually cook / are hard / 12 '' long .
    \item LIME:	comfortably / wanted / helps / tips / fit
    \item LRP (W):	are / not / 6 / hard / helps
    \item LRP (N):	are hard , not / about 6 '' between / not what I wanted / helps someone else know / wanted something I
    \item DeepLIFT (W):	are / not / 6 / hard / helps
    \item DeepLIFT (N):	are hard , not / about 6 '' between / not what I wanted / helps someone else know / wanted something I
    \item Grad-CAM-Text:	comfortably in my hand / I wanted : These / . The tips open / , not rubbery texture / Hope this helps someone
    \item DTs:	imagined . The tips

\end{itemize}
\newpage
\noindent\textbf{Top-5 counter-evidence texts}
\begin{itemize}
    \item Random (W):	texture / . / what / to / would
    \item Random (N):	this helps someone else / , not / wanted something I / and the / I did n't
    \item LIME	not / else / someone / ok / would
    \item LRP (W):	: / tips / open / in / The
    \item LRP (N):	. The tips open / : These would / in my hand and / could actually cook / I did n't realize
    \item DeepLIFT (W):	: / tips / open / in / The
    \item DeepLIFT (N):	. The tips open / : These would / in my hand and / could actually cook / I did n't realize
    \item Grad-CAM-Text:	not what I wanted / not rubbery texture like / Hope this helps someone / would be ok / The handles did n't
    \item DTs:	'd imagined . / are . I wanted / would be ok
\end{itemize}

\newpage
\noindent \underline{\textbf{Example 4}}: ArXiv Dataset, (Actual: Computer Science (CS), Predicted: Mathematics (MA)),
Predicted scores: Computer Science (0.108), Mathematics (0.552), Physics (0.340)

\vspace{0.2cm}
\noindent\fbox{%
    \parbox{0.97\linewidth}{%
``The mnesor theory is the adaptation of vectors to artificial intelligence. The scalar field is replaced by a lattice. Addition becomes idempotent and multiplication is interpreted as a selection operation. We also show that mnesors can be the foundation for a linear calculus.''
    }%
}

\vspace{0.2cm}
\noindent\textbf{Top-5 evidence texts}
\begin{itemize}
        \item Random (W):	intelligence / to / theory / is / by
    \item Random (N):	replaced by a lattice / interpreted as a / linear calculus . / show that / The mnesor
    \item LIME:	linear / a / idempotent / vectors / of
    \item LRP (W):	lattice / theory / scalar / linear / of
    \item LRP (N):	replaced by a lattice / . The scalar field / the adaptation of vectors / mnesor theory / a linear
    \item DeepLIFT (W):	lattice / theory / scalar / linear / of
    \item DeepLIFT (N):	replaced by a lattice / . The scalar field / the adaptation of vectors / mnesor theory / a linear
    \item Grad-CAM-Text:	for a linear calculus / Addition becomes idempotent and / adaptation of vectors to / replaced by a lattice / mnesor theory is the
    \item DTs:	Addition becomes idempotent and / becomes idempotent and multiplication

\end{itemize}
\newpage
\noindent\textbf{Top-5 counter-evidence texts}
\begin{itemize}
        \item Random (W):	the / We / scalar / lattice / operation
    \item Random (N):	lattice . Addition / The scalar / interpreted as a selection / for a linear calculus / . 
    \item LIME:	intelligence / scalar / field / The / lattice 
    \item LRP (W):	mnesors / interpreted / multiplication / can / foundation
    \item LRP (N):	mnesors can be the / multiplication is interpreted as / to artificial intelligence / foundation for / field is
    \item DeepLIFT (W):	interpreted / mnesors / multiplication / foundation / can 
    \item DeepLIFT (N):	mnesors can be the / multiplication is interpreted as / to artificial intelligence / foundation for / field is 
    \item Grad-CAM-Text:	. The scalar field / vectors to artificial intelligence / show that mnesors can / and multiplication is interpreted / The mnesor theory is 
    \item DTs:	vectors to artificial 

\end{itemize}

\newpage
\onecolumn
\section{Score Distributions}
This section presents the distributions of individual scores rated by human participants for each task and dataset. We do not include the random baselines in the plots to reduce the plot complexity.
\subsection{Amazon Dataset}
\begin{figure}[!h]
\centering
    \begin{subfigure}
        \centering
        \includegraphics[height=1.3in]{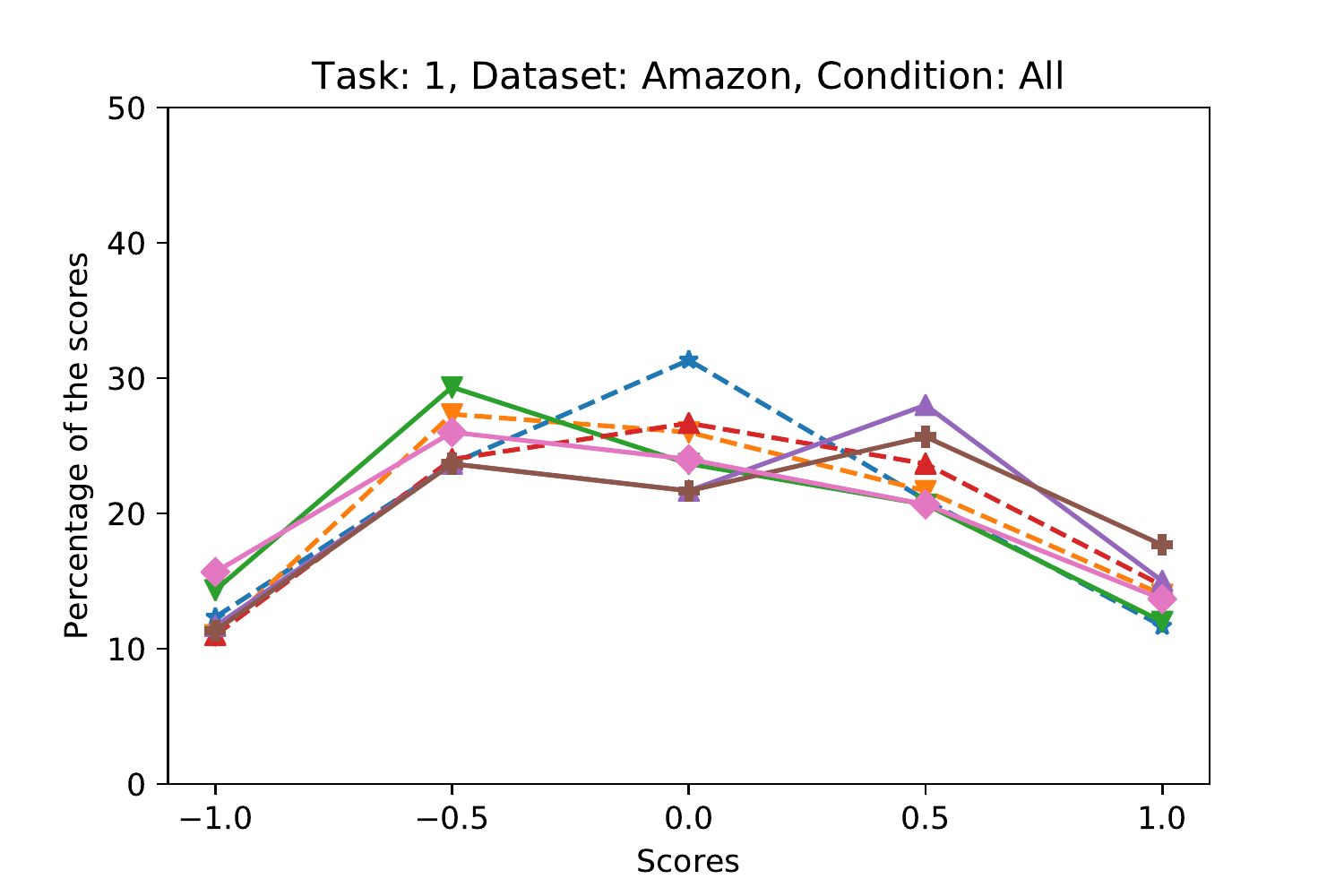}
    \end{subfigure}%
    \begin{subfigure}
        \centering
        \includegraphics[height=1.3in]{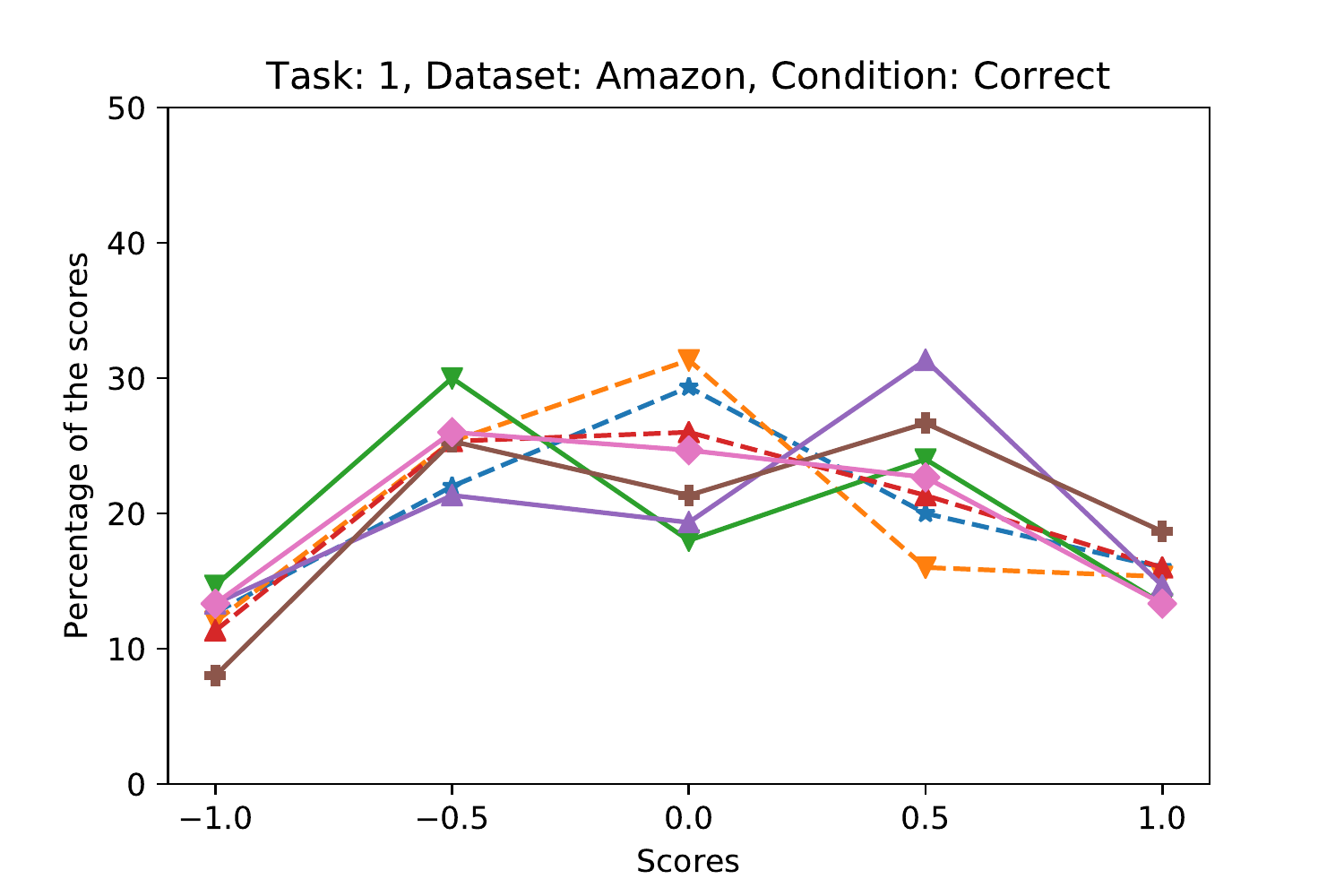}
    \end{subfigure}
    \begin{subfigure}
        \centering
        \includegraphics[height=1.3in]{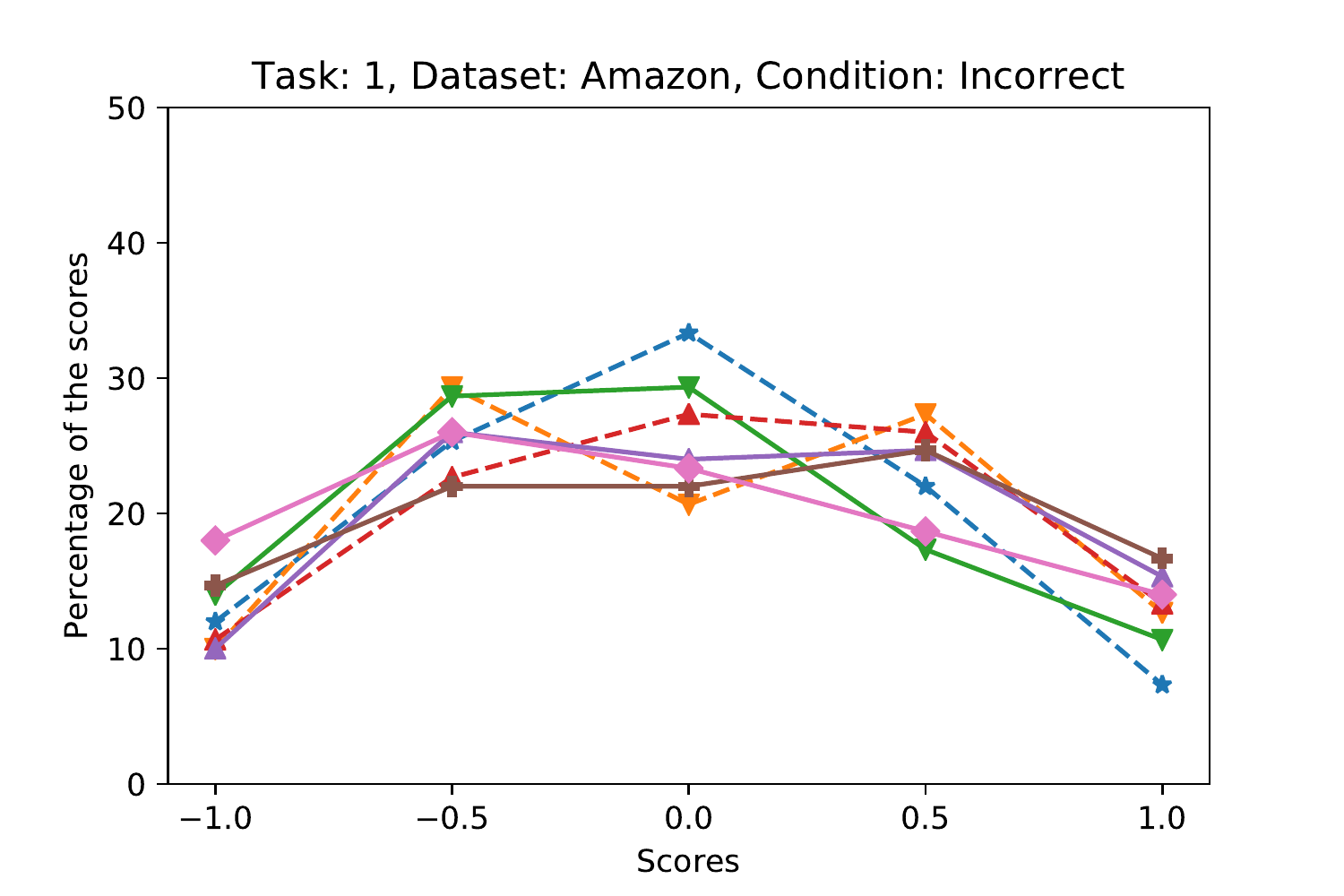}
    \end{subfigure}
    \includegraphics[width=\linewidth]{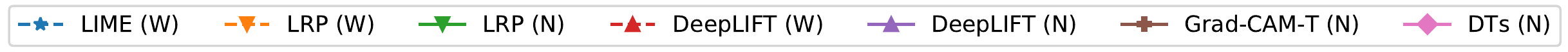}
\caption{Distributions of individual scores from task 1 of the Amazon dataset ($\mathcal{A}$, \cmark, \xmark, respectively).}\label{fig:amtask1}
\end{figure}

\begin{figure}[!h]
\centering
    \begin{subfigure}
        \centering
        \includegraphics[height=1.3in]{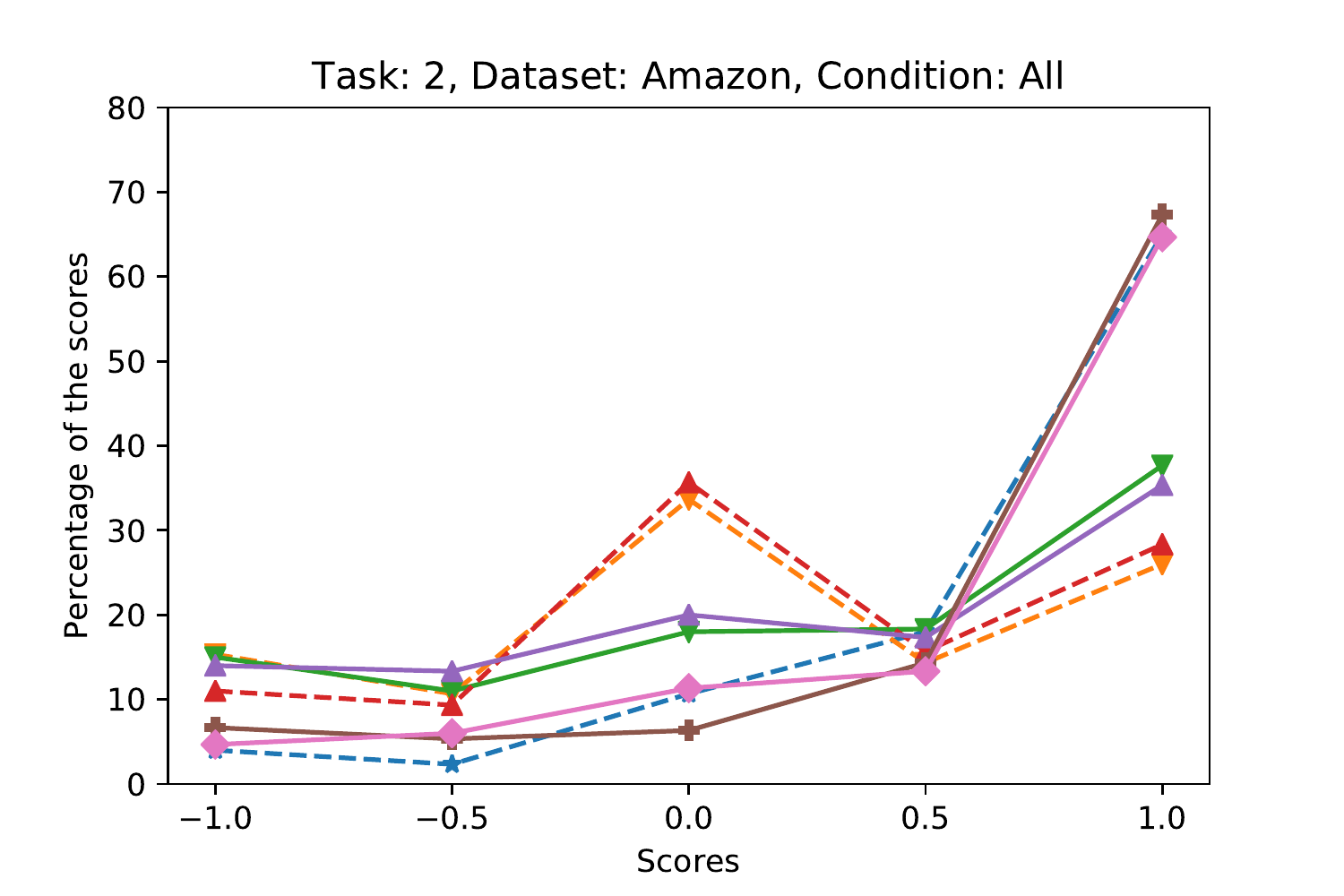}
    \end{subfigure}%
    \begin{subfigure}
        \centering
        \includegraphics[height=1.3in]{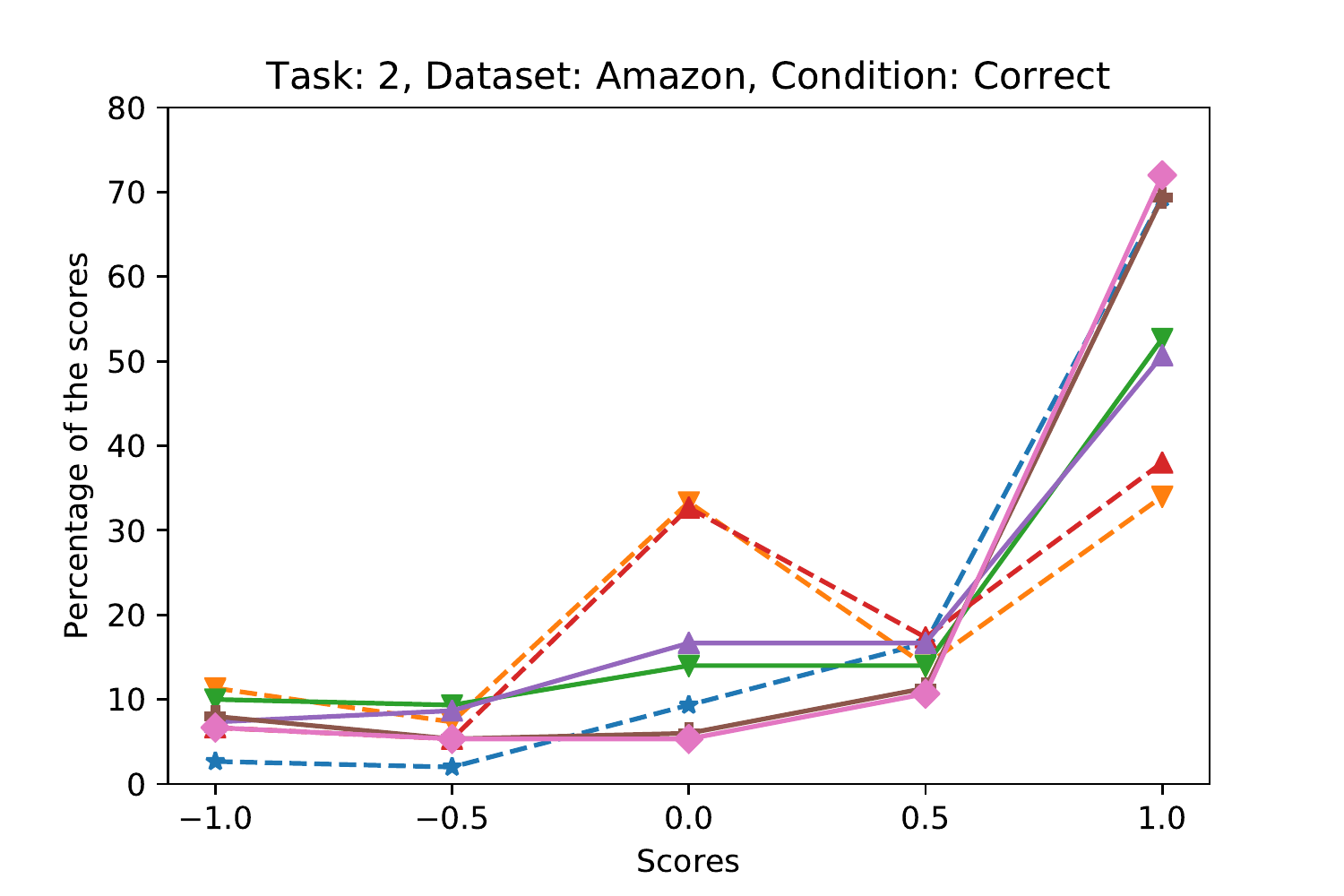}
    \end{subfigure}
    \begin{subfigure}
        \centering
        \includegraphics[height=1.3in]{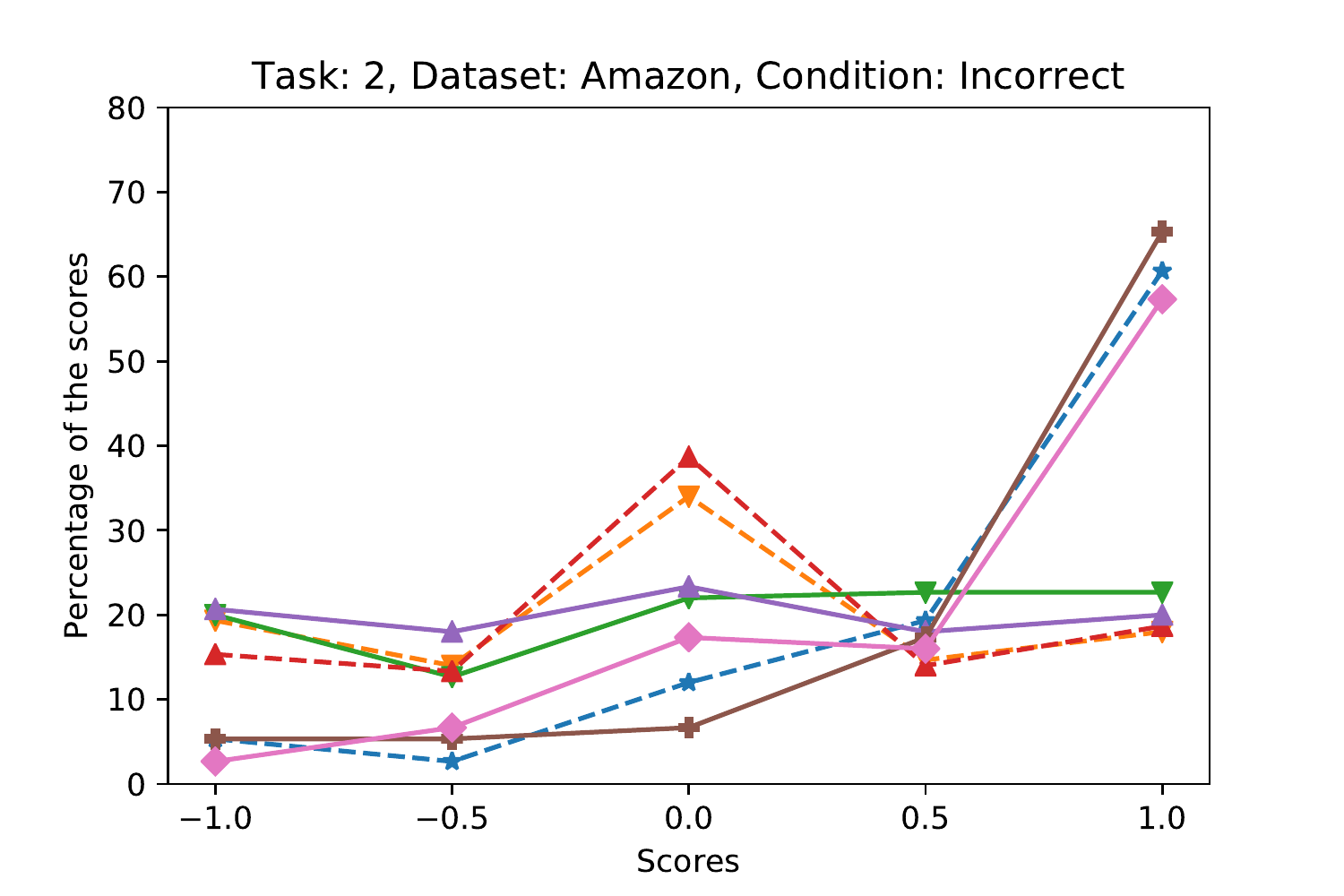}
    \end{subfigure}
    \includegraphics[width=\linewidth]{fig/legend.pdf}
\caption{Distributions of individual scores from task 2 of the Amazon dataset ($\mathcal{A}$, \cmark, \xmark, respectively).}\label{fig:amtask2}
\end{figure}

\begin{figure}[!h]
\centering
    \begin{subfigure}
        \centering
        \includegraphics[height=1.3in]{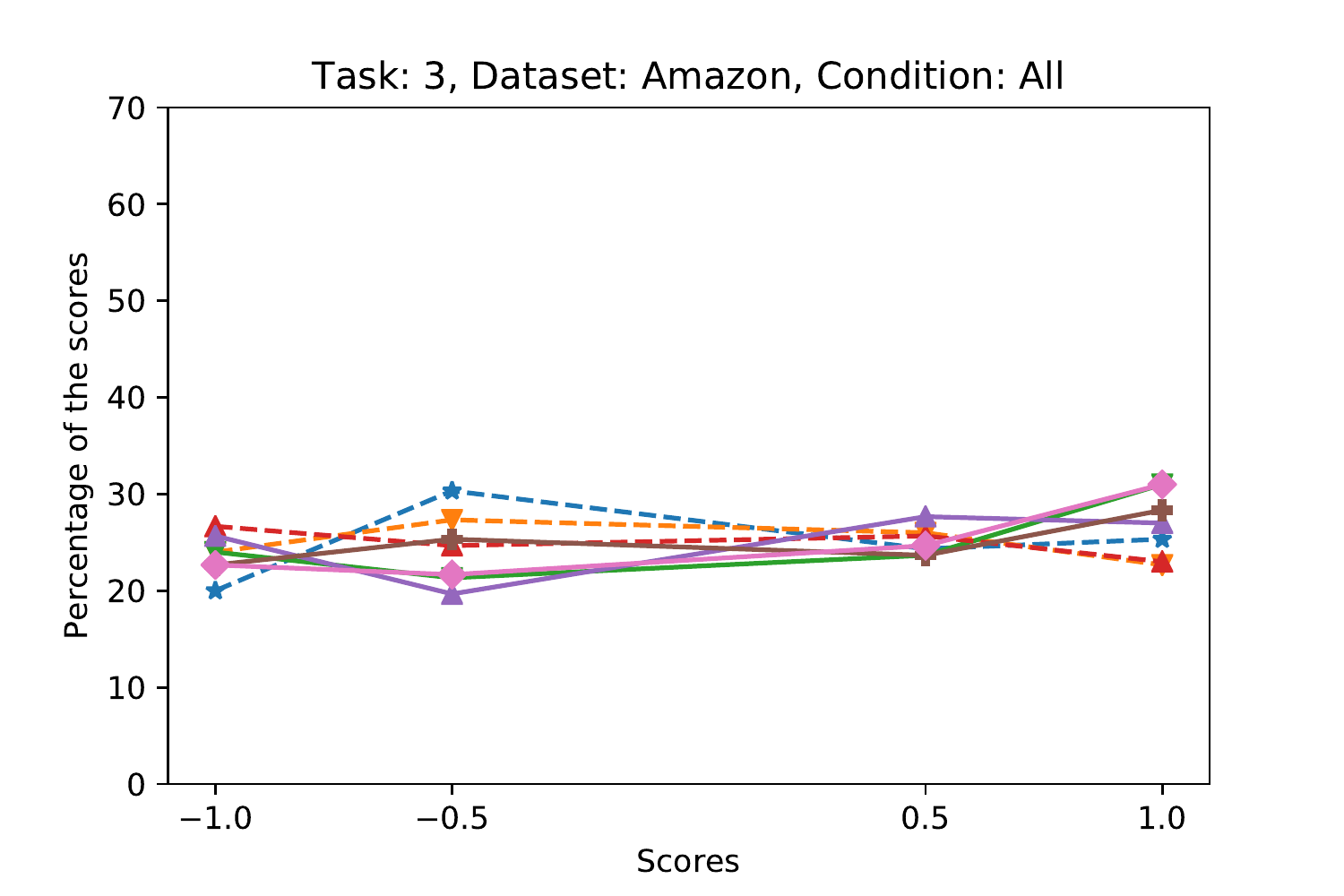}
    \end{subfigure}%
    \begin{subfigure}
        \centering
        \includegraphics[height=1.3in]{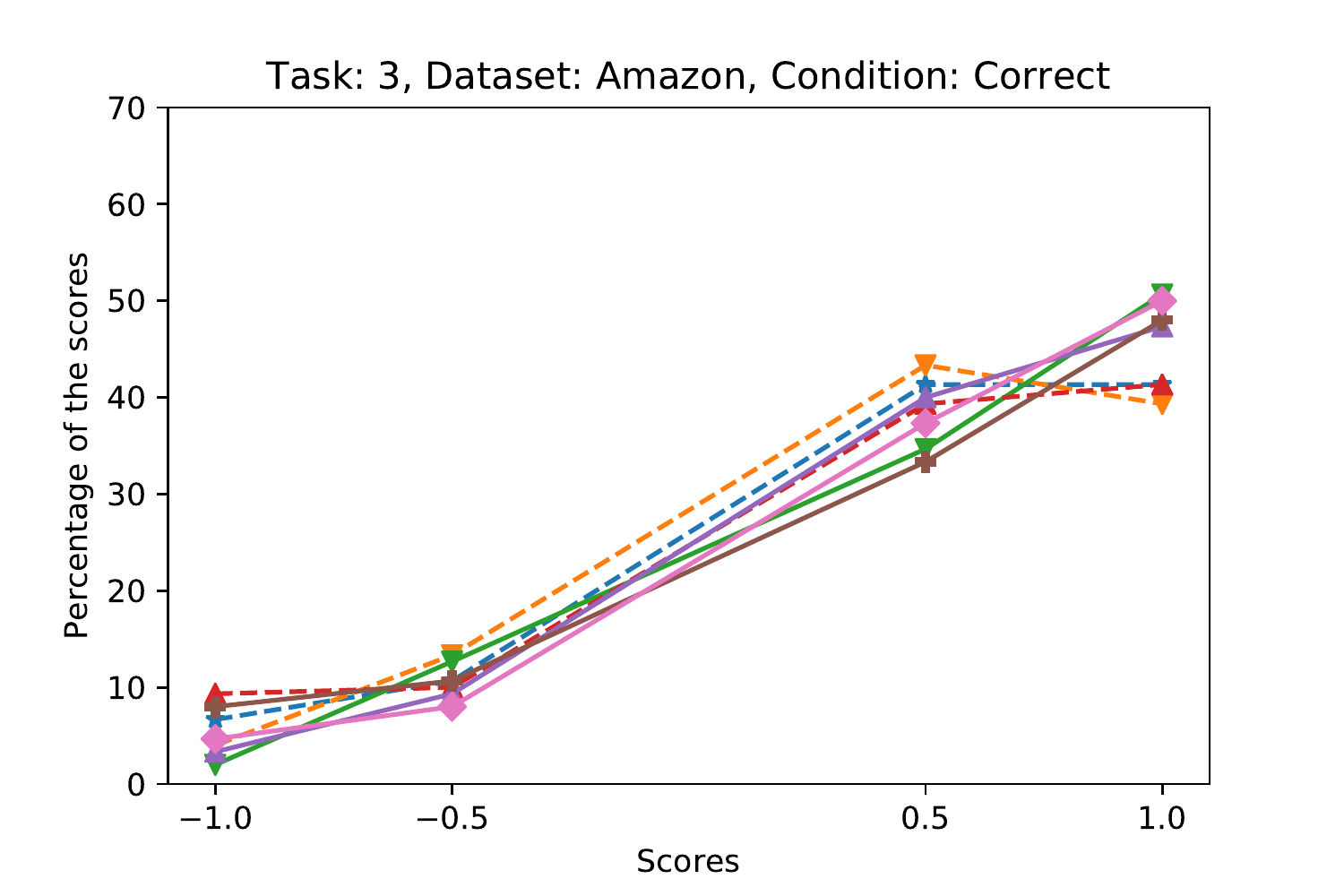}
    \end{subfigure}
    \begin{subfigure}
        \centering
        \includegraphics[height=1.3in]{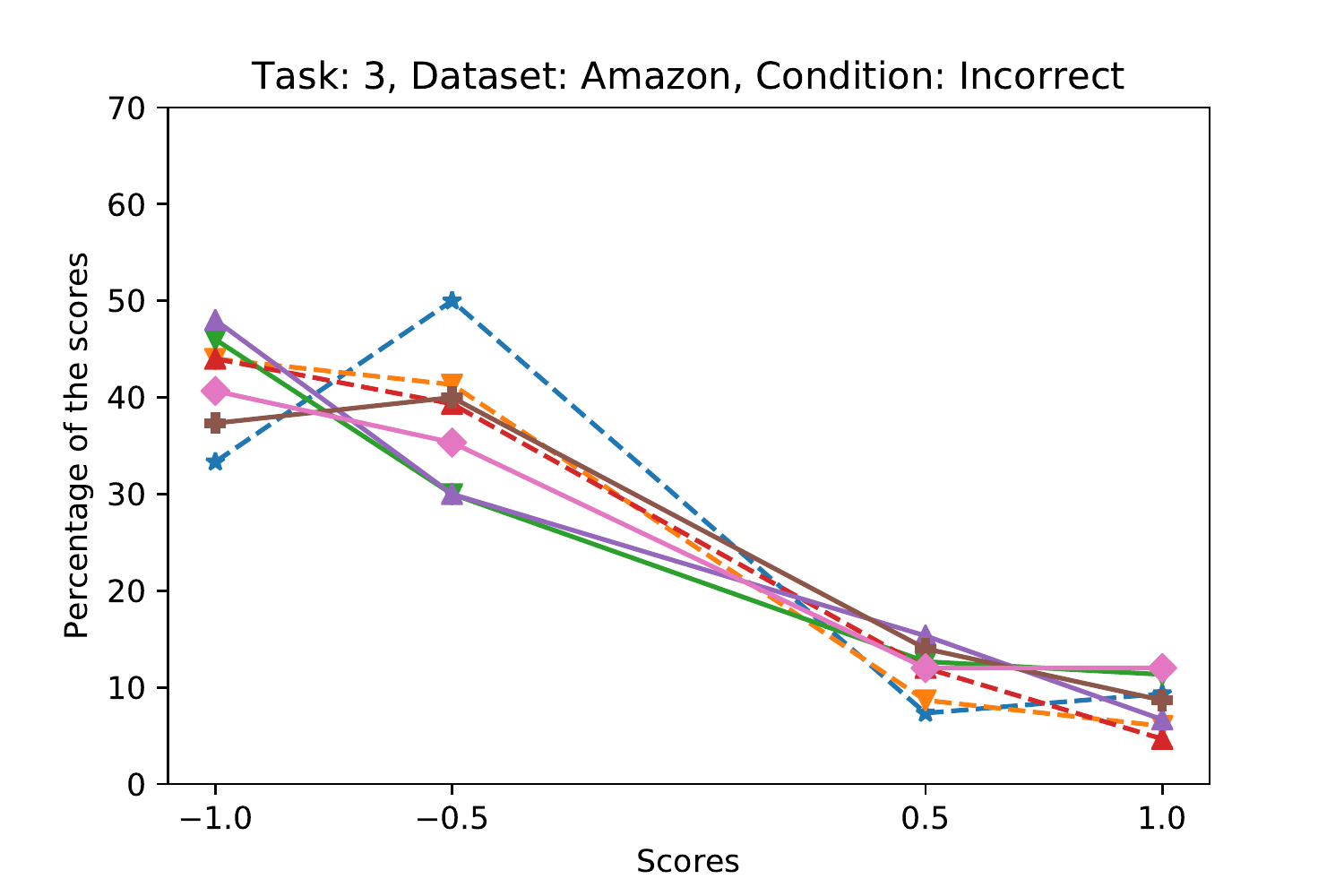}
    \end{subfigure}
    \includegraphics[width=\linewidth]{fig/legend.pdf}
\caption{Distributions of individual scores from task 3 of the Amazon dataset ($\mathcal{A}$, \cmark, \xmark, respectively).}\label{fig:amtask3}
\end{figure}

\newpage
\subsection{ArXiv Dataset}

\begin{figure}[!h]
\centering
    \begin{subfigure}
        \centering
        \includegraphics[height=1.3in]{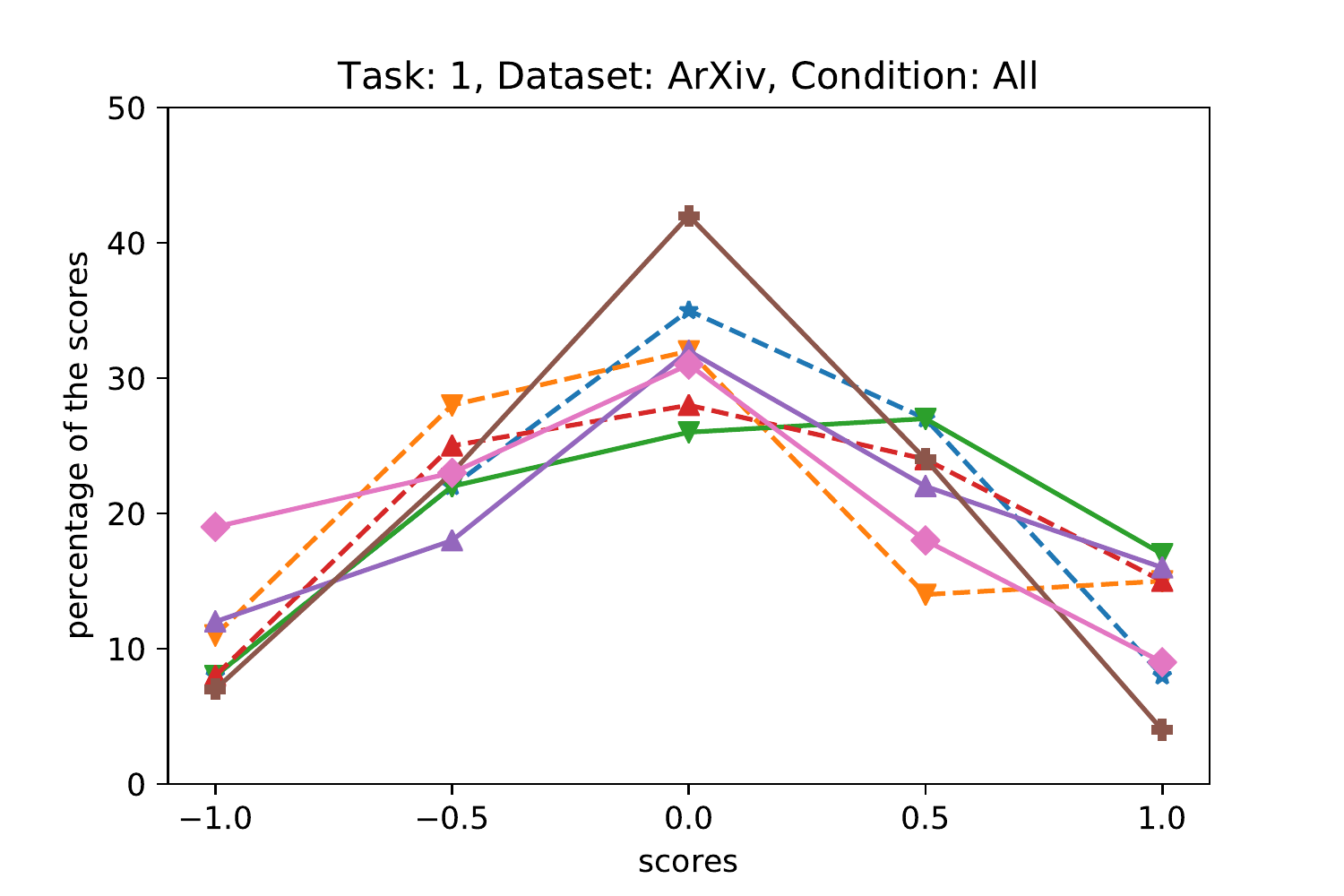}
    \end{subfigure}%
    \begin{subfigure}
        \centering
        \includegraphics[height=1.3in]{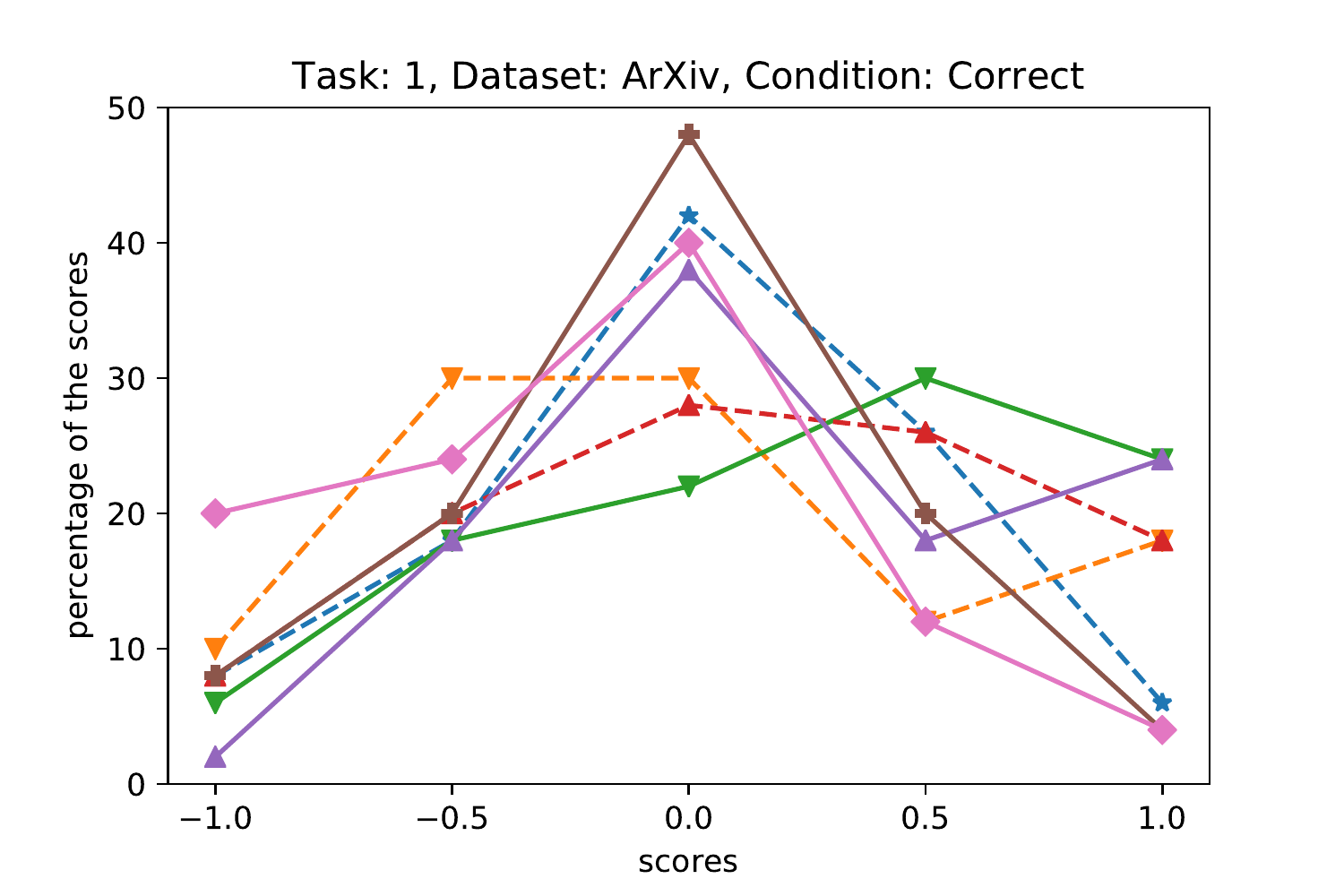}
    \end{subfigure}
    \begin{subfigure}
        \centering
        \includegraphics[height=1.3in]{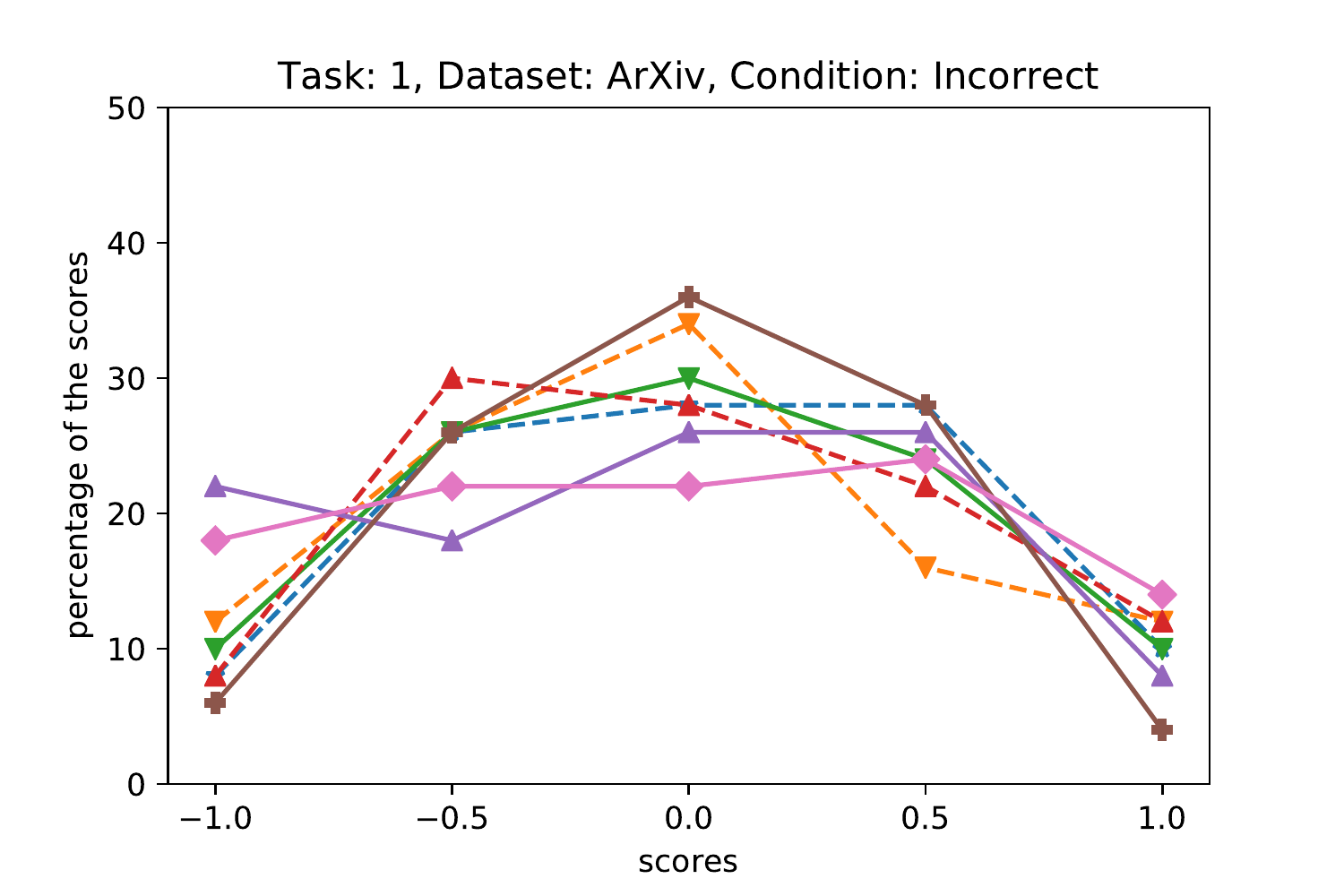}
    \end{subfigure}
    \includegraphics[width=\linewidth]{fig/legend.pdf}
\caption{Distributions of individual scores from task 1 of the ArXiv dataset ($\mathcal{A}$, \cmark, \xmark, respectively).}\label{fig:artask1}
\end{figure}

\begin{figure}[!h]
\centering
    \begin{subfigure}
        \centering
        \includegraphics[height=1.3in]{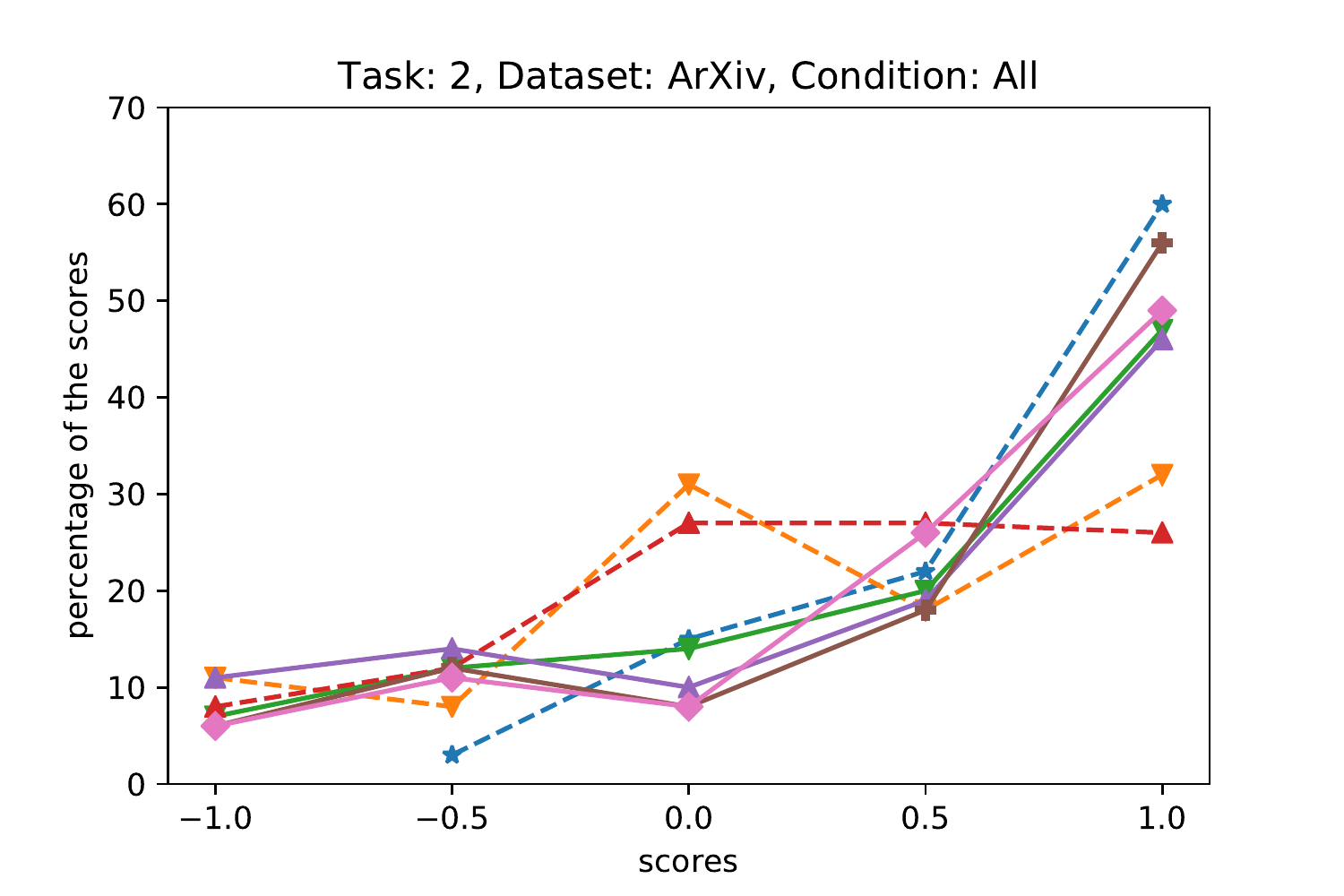}
    \end{subfigure}%
    \begin{subfigure}
        \centering
        \includegraphics[height=1.3in]{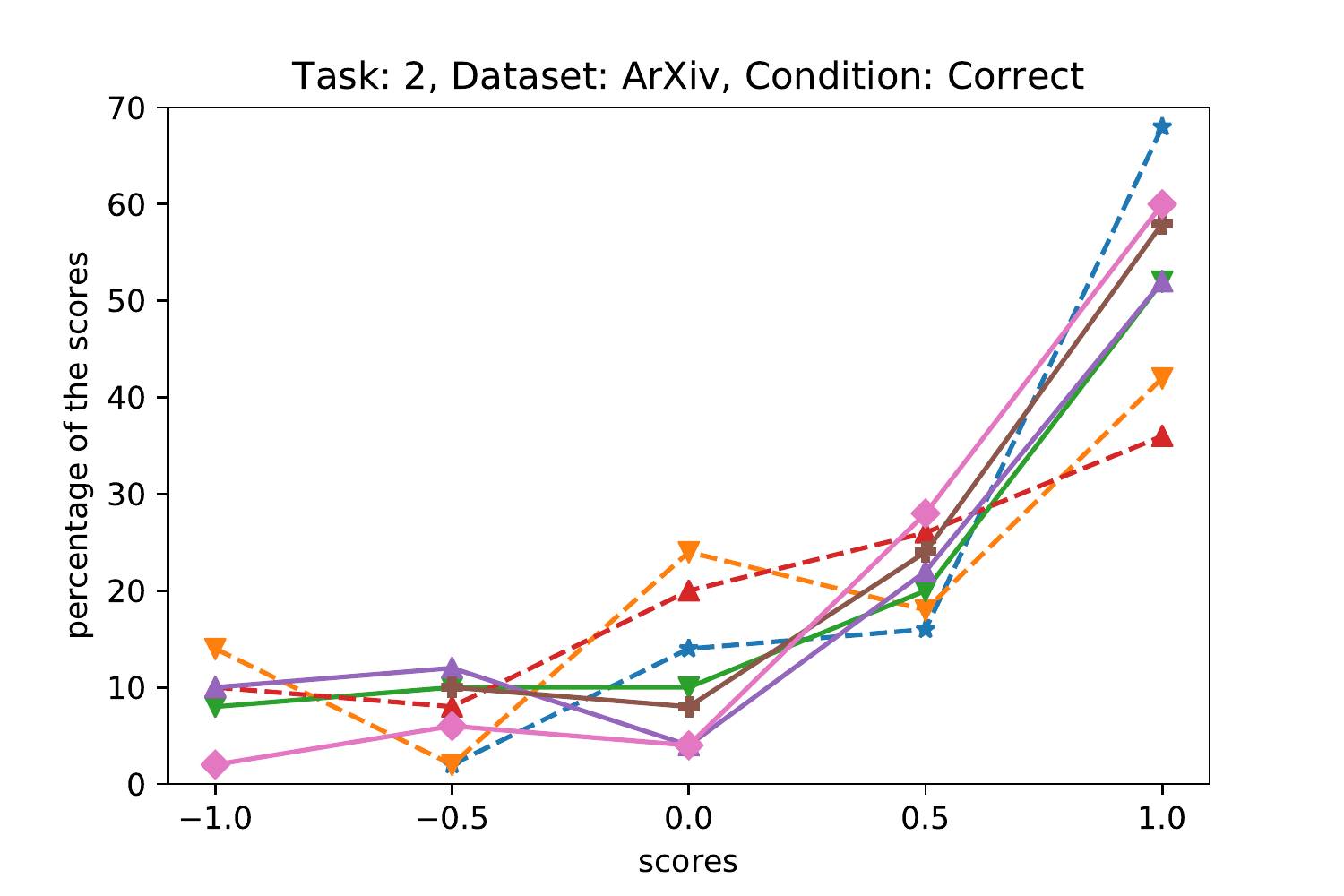}
    \end{subfigure}
    \begin{subfigure}
        \centering
        \includegraphics[height=1.3in]{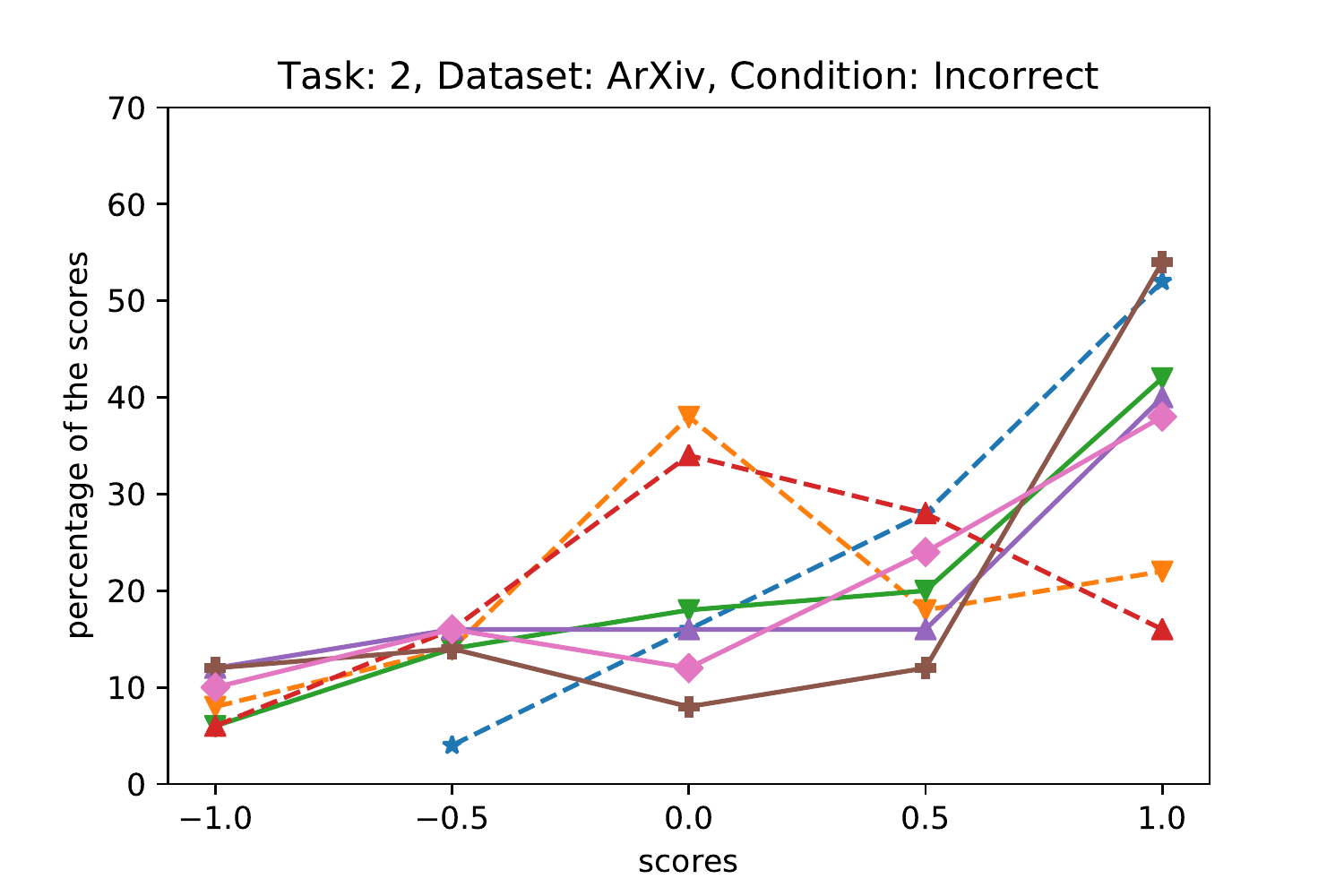}
    \end{subfigure}
    \includegraphics[width=\linewidth]{fig/legend.pdf}
\caption{Distributions of individual scores from task 2 of the ArXiv dataset ($\mathcal{A}$, \cmark, \xmark, respectively).}\label{fig:artask2}
\end{figure}

\begin{figure}[!h]
\centering
    \begin{subfigure}
        \centering
        \includegraphics[height=1.3in]{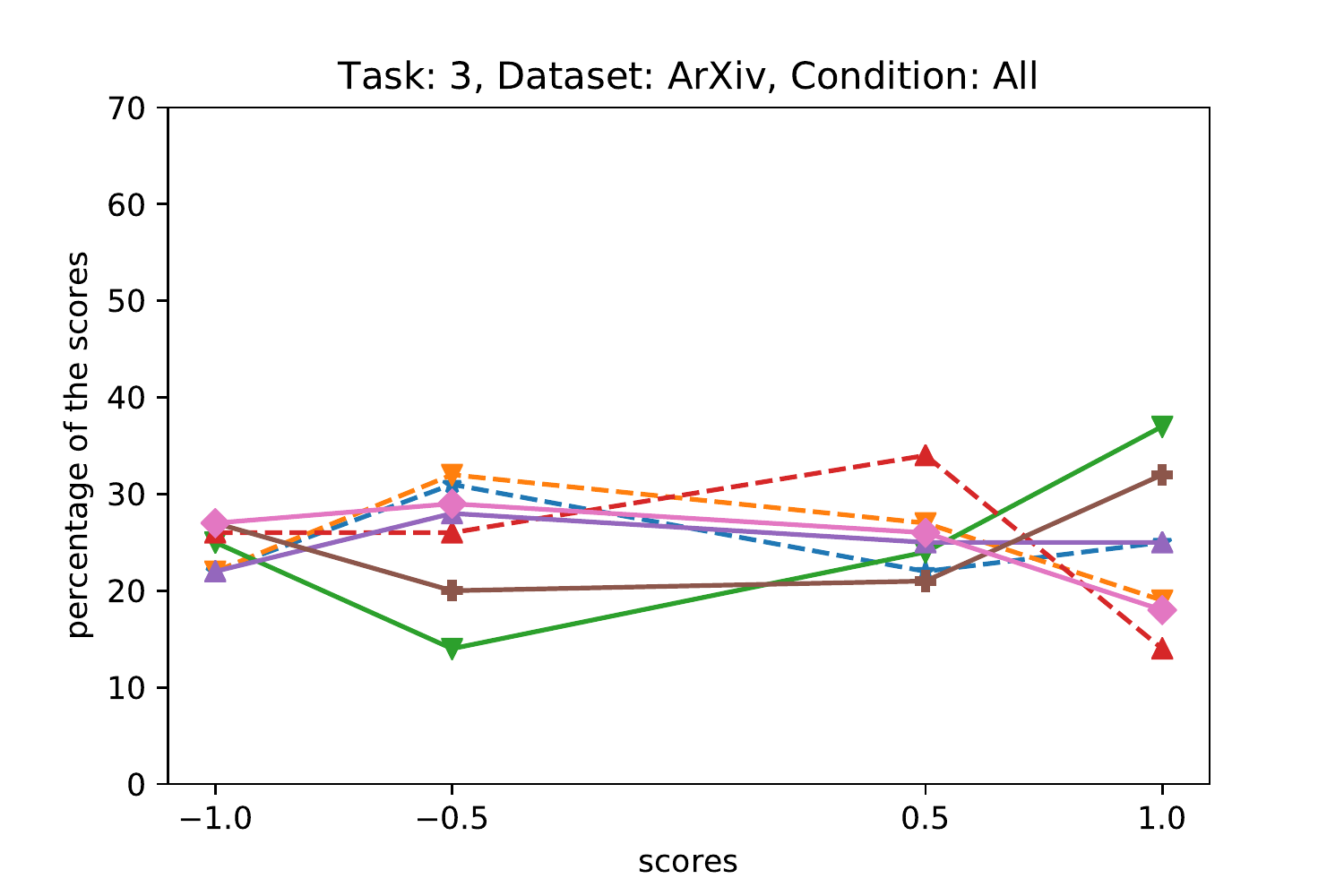}
    \end{subfigure}%
    \begin{subfigure}
        \centering
        \includegraphics[height=1.3in]{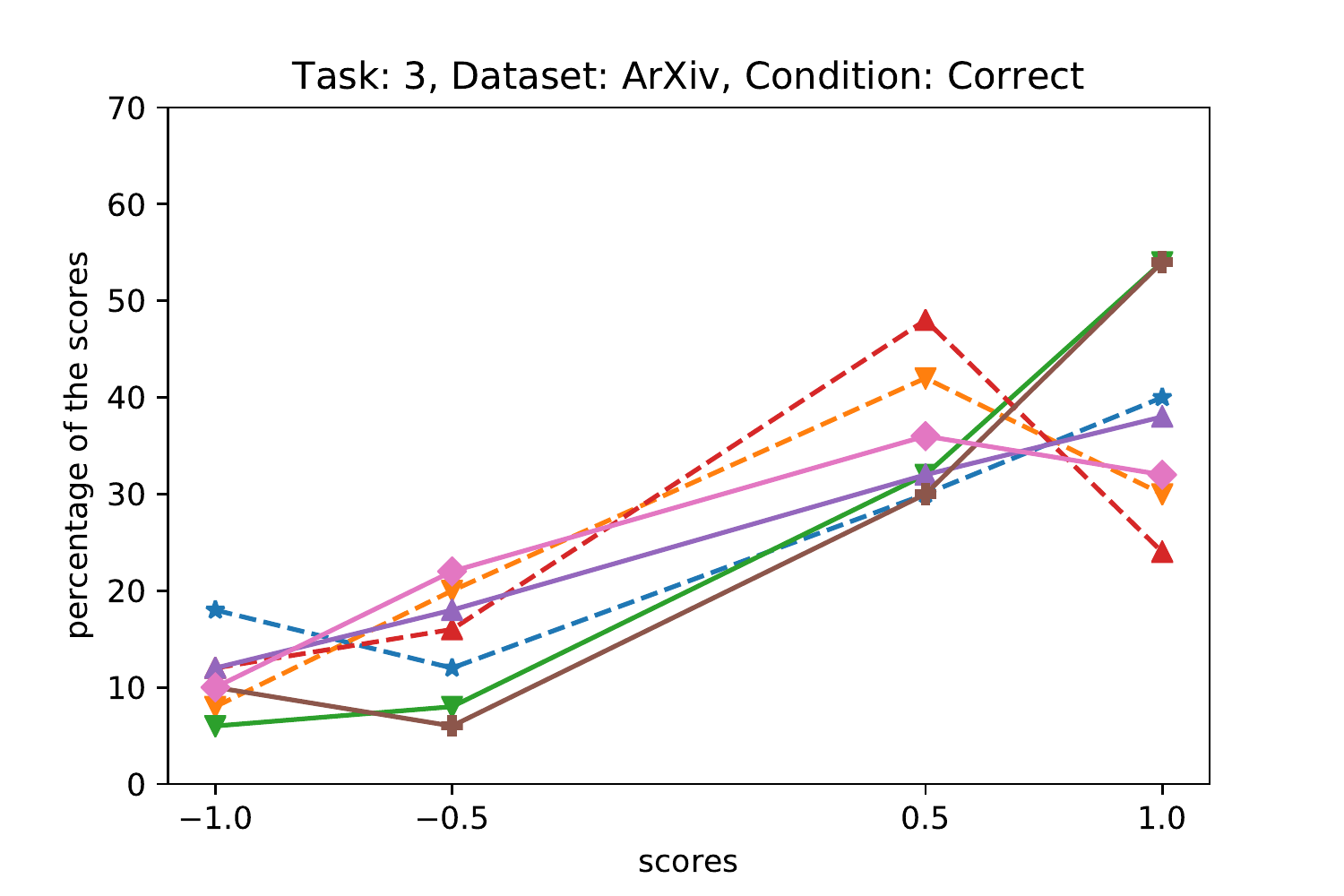}
    \end{subfigure}
    \begin{subfigure}
        \centering
        \includegraphics[height=1.3in]{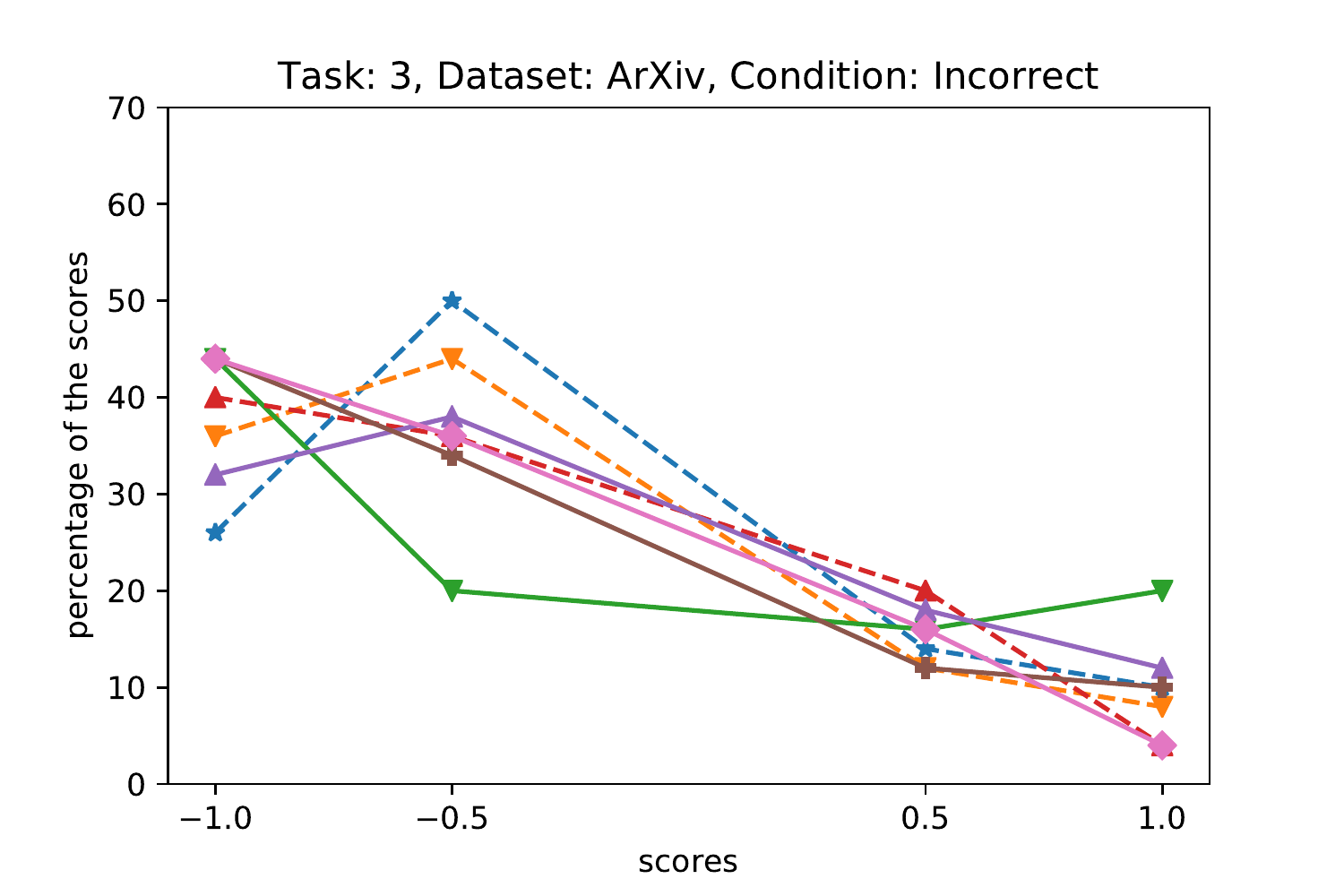}
    \end{subfigure}
    \includegraphics[width=\linewidth]{fig/legend.pdf}
\caption{Distributions of individual scores from task 3 of the ArXiv dataset ($\mathcal{A}$, \cmark, \xmark, respectively).}\label{fig:artask3}
\end{figure}
\end{document}